\pdfoutput=1
\documentclass[journal]{IEEEtran}

\usepackage[utf8]{inputenc}
\usepackage[T1]{fontenc}
\usepackage{amsmath,amsfonts,amssymb}
\usepackage{array}
\usepackage[caption=false,font=footnotesize]{subfig}
\usepackage{stfloats}
\usepackage{url}
\usepackage{graphicx}
\usepackage{xcolor}
\usepackage{cite}
\usepackage{tikz}
\usetikzlibrary{arrows.meta,calc,decorations.pathreplacing,positioning}

\graphicspath{{figures/}}

\begin{document}

\title{Race to Recovery: Designing Advanced Safety Systems for Autonomous Driving via Learning Autonomous Racing}

\author{Teodor~Ilie, Kleber~Cabral, and~Sidney~Givigi
\thanks{This work has been submitted to the IEEE for possible publication. Copyright may be transferred without notice, after which this version may no longer be accessible. (Corresponding author: Teodor Ilie.)}%
\thanks{We acknowledge the support of the Natural Sciences and Engineering
Research Council of Canada (NSERC). Nous remercions le Conseil de recherches en
sciences naturelles et en g\'enie du Canada (CRSNG) de son soutien. T. Ilie was
supported by an NSERC Canada Graduate Scholarship~--~Master's and by a Vector
Scholarship in Artificial Intelligence. The authors declare no conflicts of
interest.}%
\thanks{The authors are with the School of Computing, Queen's University,
Kingston, ON K7L 2N8, Canada (e-mail: teodor.ilie@queensu.ca;
kleber.cabral@queensu.ca; sidney.givigi@queensu.ca).}}

\markboth{Preprint submitted to IEEE Transactions on Intelligent Vehicles}%
{Ilie \MakeLowercase{\textit{et al.}}: Race to Recovery: Designing Advanced Safety Systems for Autonomous Driving via Learning Autonomous Racing}

\maketitle

\begin{abstract}
Vehicles that lose control at the handling limit through icing, hydroplaning, or an abrupt evasive maneuver fall outside the operating envelope of production safety systems such as electronic stability control, which focus on preventing instability rather than recovering from it. Expert drivers regain authority in these conditions by drifting. Existing autonomous recovery approaches are validated on idealized scenarios, define ``recovered'' states empirically and inconsistently, or rely on offline-planned trajectories rather than acting online. We address these gaps by first training a racing policy with reinforcement learning that learns drifting as a time-optimal maneuver, then adapting it to recovery through continued training with a minimum-time reward. We define recovery relative to a phase-plane-grounded stable equilibrium and introduce Gym-Khana for controller training and evaluation. Across 294 initial loss-of-control states, the transferred policy improves recovery by $186.8\%$ over a Stanley path tracker and outperforms both a nonlinear model predictive controller and a recovery-only policy trained from scratch, while also recovering faster on states the baseline can already handle. An ablation shows that racing pretraining, rather than broader initial-state exposure alone, drives this gain. Deployed on a physical 1:10-scale F1TENTH vehicle using system identification, actuator-lag modelling, and domain randomization, the transferred policy outperforms Stanley on every tested loss-of-control state, including recovering $50\%$ of the time from a hard state the baseline never recovers from.
\end{abstract}

\begin{IEEEkeywords}
Reinforcement learning, transfer learning, vehicle dynamics, vehicle safety.
\end{IEEEkeywords}

\section{Introduction}
\label{sec:intro}

Usually, driving takes place in the linear tire regime, where manual inputs or conventional controllers perform well. Controlling a vehicle safely at the limit of handling remains an open challenge. Events that could cause loss of control, such as driving on ice, high-speed hydroplaning, or a sudden evasive maneuver, are rare but dangerous, and excessive speed alone is a factor in roughly a third of all traffic fatalities~\cite{TrafficSafetyFacts2024}. They are also the events an advanced driver assistance system (ADAS) is least able to help with. Production systems such as electronic stability control (ESC) are designed to keep the vehicle inside a conservative, linear-tire handling envelope, braking individual wheels to suppress the onset of tire saturation. That design is well suited to preventing a loss of control, but not to act once one has occurred: beyond this stable envelope, suppressing instability is no longer the same problem as recovering from it.

Expert rally and drift drivers, by contrast, routinely operate at the limit by ``drifting'', where they intentionally saturate the tires to sustain large sideslip angles. Saturation decouples the direction the vehicle points from the direction it travels, giving the driver direct lateral control through two inputs, steering and throttle, rather than one, enabling more agile vehicle control~\cite{hindiyehDrifting}. This is the authority an expert exploits when control is lost, while current production safety systems cannot. Recent research has begun to reproduce this capability autonomously in the service of active safety, using drift-based control to regain control of the vehicle in scenarios where conventional safety systems fall short~\cite{liPlanningControlDriftingbased2023a, stanoEnhancedActiveSafety2024a}. We call this task \textit{recovery}: returning an out-of-control vehicle to a stable driving state while keeping it on its intended path. We aim to develop a controller that recovers from the broadest possible set of loss-of-control states and recovers in the least time while remaining within the track. Our motivation is to endow an ADAS with the recovery skill of an expert driver.

Existing autonomous approaches to recovery, however, leave three elements unresolved. Each addresses a single narrow, idealized scenario from a fixed initial condition, so nothing is established about the range of loss-of-control states from which a controller can recover~\cite{yinSelflearningDriftControl2020a, leeLearningbasedAutonomousOversteer2025a}. Each defines the ``recovered'' state empirically, and differently, so no two results are comparable. And the one method validated on a real vehicle relies on a trajectory planned offline in advance~\cite{liPlanningControlDriftingbased2023a}, whereas an intervention must act online, from whatever state the vehicle is in.

A recovery controller that closes these gaps must therefore act online, without a precomputed plan, across a broad domain of loss-of-control states in the saturated-tire regime. That is the regime of racing and drifting at the limit of handling, where reinforcement learning (RL) has been shown to outperform state-of-the-art model predictive control (MPC)~\cite{songReachingLimitAutonomous2023a, czechmanowskiLearningRacingPolicies2025b}.
We therefore derive the controller with RL, focusing on the control task in particular, assuming knowledge of the vehicle state, and leave perception, state estimation, and planning out of our scope. The open question is how to set up training.

Our central insight is that the answer lies in racing. The recovery task has much in common with drifting: both are defined primarily by a large sideslip angle and yaw rate, but where a racing driver produces the drift state intentionally, a loss of control is an unwanted drift that can lead to a collision. A policy that has learned to race by drifting has, in the process, learned to manage saturated tires and large sideslip angles across the whole regime, which is precisely the competency that recovery requires. We therefore investigate the following hypothesis: a policy first trained to race with drifting, and then adapted to recovery, recovers control more reliably than a policy trained on recovery alone, and more reliably than classical control baselines. To test it, we train a racing policy in Gym-Khana, a simulator we built for prototyping vehicle control at the limit, adapt it to recovery by continuing training under a minimum-time recovery reward, and deploy the result on a 1:10-scale F1TENTH vehicle~\cite{okellyF110OpenSource2019}.

In summary, our contributions are:
\begin{itemize}
    \item A racing-to-recovery transfer learning method, which builds a recovery controller by pretraining a policy to race with drifting and then adapting it to recovery. Over a grid of $294$ initial states spanning conditions more extreme than any seen in training, the transferred policy recovers $186.8\%$ more than a Stanley path tracker, and outperforms both a model predictive controller and a recovery-only policy trained from scratch.
    Additionally, we demonstrate that even non-specialized racing policies recover competently with no recovery-specific training (Sections~\ref{sec:transfer} and~\ref{sec:ablation}).
    \item Gym-Khana, a simulator for prototyping vehicle control at the limit of handling, built around the drift-capable single-track model used here, on which both the racing and recovery policies are trained (Section~\ref{sec:method}). It is released open-source, with online documentation.\footnote{\raggedright Gym-Khana simulator: \url{https://github.com/TeoIlie/Gym-Khana}; documentation: \url{https://gym-khana.readthedocs.io}}
    \item A rigorous, phase-plane-grounded definition of vehicle recovery that formalizes the ``recovered'' state as a tolerance region about a stable equilibrium, together with a set of experimental recovery metrics that make recovery controllers comparable over a common domain of loss-of-control states (Sections~\ref{sec:recovery_task} and~\ref{sec:baselines_metrics}).
    \item Validation on a physical F1TENTH vehicle, enabled by a sim-to-real pipeline of system identification, actuator-lag modelling, and domain randomization. Over $179$ runs from three real loss-of-control initial states of increasing difficulty, the transferred policy outperforms the Stanley baseline on every state (Section~\ref{sec:hardware_validation}).
\end{itemize}

\section{Related Work}
\label{sec:related}
This section reviews prior work on vehicle control at and beyond the limit of handling, with particular emphasis on RL for high-performance racing and drifting. We first examine the progression from model-based optimal control, particularly model predictive control, to learning-based approaches that have demonstrated advantages at the handling limit. We then review emerging applications of drifting for active safety and recovery, focusing on how these methods define and achieve recovery from loss-of-control scenarios.

\subsection{Control at the Limit of Handling: From MPC to RL}
\label{sec:related_limit}

Control at the limit of handling has historically been approached with optimal control, most commonly MPC. The model predictive contouring control (MPCC) formulation, which maximizes progress along a reference centreline subject to track constraints, has been applied successfully to both autonomous drone and vehicle racing~\cite{linigerOptimizationBasedAutonomousRacing2015, romeroModelPredictiveContouring2022}, and model-based methods have achieved controlled drifting on full-scale vehicles~\cite{hindiyehDrifting, petersonExploitingLinearStructure2023, weberModelingControlDynamic2024, djeumouOneModelDrift2025}. The cost of the paradigm is a nonconvex optimization over combined-slip tire forces at every step of the control loop, which bounds model fidelity, horizon length and control rate even with efficient modelling and solver tooling~\cite{anderssonCasADiSoftwareFramework2019, verschuerenAcadosModularOpensource2021}, and a cost function that must be hand-tuned.

RL has since overtaken these methods at the limit. In the highly realistic Gran Turismo racing simulator, a policy was the first to achieve superhuman lap times, learning from a reward of centreline progress and a wall-collision penalty---the MPCC objective, optimized directly rather than through a planner~\cite{fuchsSuperHumanPerformanceGran2021}. Two years later, ``Swift'' became the first system to outperform champion human pilots at drone racing, using onboard sensing alone~\cite{kaufmannChampionlevelDroneRacing2023a}. A companion study compared RL against optimal control on the same task, found RL to outperform it, and identified the reason: the fundamental advantage of RL is ``not that it optimizes its objective better but that it optimizes a better objective''~\cite{songReachingLimitAutonomous2023a}---it optimizes the objective of interest directly, rather than through a planner and a limited intermediate representation.

The same progression has played out in drifting. Early work learned drift control in simulation~\cite{cutlerAutonomousDriftingUsing2016, caiHighspeedAutonomousDrifting2020a} and transferred a policy rewarded for path tracking and sideslip onto a 1:10-scale car using domain randomization on tire friction~\cite{dombergDeepDriftingAutonomous2022}; more recent work has reached full-scale vehicles and increasingly fuses learning with physics-based structure~\cite{tothSimtoRealApplicationReinforcement2024, djeumouReferenceFreeFormulaDrift2025}. Czechmanowski et al.~\cite{czechmanowskiLearningRacingPolicies2025b} brought the two threads together with a policy trained on a single-track model with combined-slip Pacejka tires and deployed zero-shot on a 1:8-scale car, the first controller to outperform both expert human drivers and a state-of-the-art MPC at racing. Drifting emerged there \textit{organically}, as the fastest way around the track, from nothing more than a progress reward and an out-of-bounds penalty. Sim-to-real transfer combined actuator-lag modelling, learning-based system identification and targeted domain randomization---the three ingredients we build on in Section~\ref{sec:sim2real-transfer}.

Across racing and drifting, then, RL is the strongest demonstrated approach to control at the limit of handling, outperforming both expert humans and state-of-the-art MPC. We adopt it on that evidence, and scope our own baseline claim accordingly in Section~\ref{sec:baselines_metrics}.

\subsection{Drifting for Active Safety and Recovery}
\label{sec:related_safety}

An emerging body of work explores safety applications of drift-like maneuvers. Yin et al.~\cite{yinSelflearningDriftControl2020a} applied deep RL to stabilizing yaw motion following a high-speed rear-end collision in simulation, improving over a baseline of no stabilization but with no comparison against alternative controllers; an episode counted as safely completed when the vehicle came to a stop. Li et al.~\cite{liPlanningControlDriftingbased2023a} planned a drift-based U-turn to prevent a head-on collision with an obstacle and demonstrated it on a real vehicle, where it recovered in conditions that defeat automatic emergency braking and steering; its trajectory is computationally intensive enough to require offline pre-planning, so the method cannot run in real time. A nonlinear MPC has been used for path tracking with drifting to avoid a collision at a T-junction, outperforming an anti-lock braking baseline and a feedback-feedforward steering controller~\cite{stanoEnhancedActiveSafety2024a, kapaniaDesignFeedbackfeedforwardSteering2015}. Most recently, Lee and Cong~\cite{leeLearningbasedAutonomousOversteer2025a}\ learned a policy from sub-optimal demonstrations that recovers from excessive oversteer while avoiding obstacles, outperforming both pure RL and behaviour cloning. Loss of control is induced by a virtual kick plate, and recovery is defined there as maintaining less than $1^\circ$ of sideslip for 100 timesteps~\cite{leeLearningbasedAutonomousOversteer2025a}.

Together these works exhibit the three gaps that our work addresses, set out in Section~\ref{sec:intro}: highly specific task constructions lacking general applicability, a recovered state defined differently and empirically by each, and one real-vehicle validation that depends on an offline plan.

\section{Problem Formulation}
\label{sec:problem}

Our goal is to solve the control task of returning a vehicle that has lost control while tracking a straight reference path, such as a driving lane, to a safe operating regime. Formulating this task requires two components: a vehicle and tire model, which we describe in Section~\ref{sec:model}, and a rigorous definition of the recovered state, described in Section~\ref{sec:recovery_task}.

\subsection{Vehicle and Tire Model}
\label{sec:model}

The recovery task requires a dynamics model that represents the vehicle at a suitable level of complexity. The model must be expressive enough to capture the tire dynamics of a vehicle operating beyond the linear handling regime, while remaining simple enough to stay interpretable and to be identified from measurements of a physical vehicle, as we do in Section~\ref{sec:sim2real-transfer}.

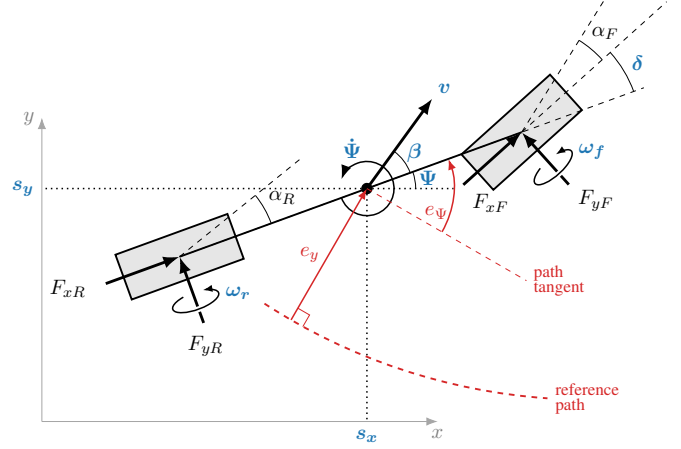
\begin{figure}[t]
  \centering
  \resizebox{\columnwidth}{!}{\fontsize{11}{13.2}\selectfont\definecolor{thesisred}{HTML}{D62728}%
\definecolor{stdblue}{HTML}{1F77B4}%
\tikzset{
  force/.style    = {-{Latex[length=3mm]}, line width=1.5pt},
  dashline/.style = {dashed, line width=0.5pt},
  projline/.style = {dotted, line width=0.8pt},
  stdstate/.style = {text=stdblue, font=\boldmath},
  axisline/.style = {-{Latex[length=2mm]}, draw=black!35, text=black!50},
  tire/.style     = {fill=black!10, line width=1pt},
  wheelspin/.pic = {
    \draw[white, line width=3pt] (245:{\spinrx} and {\spinry})
        arc[start angle=245, end angle=295, x radius=\spinrx, y radius=\spinry];
    \draw[-{Latex[length=1.8mm]}, thick] (115:{\spinrx} and {\spinry})
        arc[start angle=115, end angle=425, x radius=\spinrx, y radius=\spinry];
  },
}%
\begin{tikzpicture}
  \def\stang{22}
  \def\slip{34}
  \def\yaw{20}
  \def\fxlen{1.45}
  \def\fylen{1.3}
  \def\spinat{-0.87}
  \def\spinrx{0.45}
  \def\spinry{0.16}
  \def\axx{-6.0}
  \def\axy{-4.3}
  \def\pathang{-30}
  \def\ey{2.8}
  \def\pathR{11}

  \draw[axisline] (\axx,\axy) -- (\axx,1.3) node[left]{$y$};
  \draw[axisline] (\axx,\axy) -- (1.3,\axy) node[below]{$x$};

  \draw[projline] (0,0) -- (0,\axy) node[below, stdstate]{$s_x$};
  \draw[projline] (0,0) -- (\axx,0) node[left, stdstate]{$s_y$};

  \draw[projline] (0,0) -- (1.6,0);

  \begin{scope}[rotate=\yaw]
    \coordinate (com)   at (0,0);
    \coordinate (rear)  at (-3.68,0);
    \coordinate (front) at (3.04,0);

    \draw[tire] ($(rear)+(-1.1,-0.44)$) rectangle ($(rear)+(1.1,0.44)$);
    \begin{scope}[shift={(front)}, rotate=\stang]
      \draw[tire] (-1.1,-0.44) rectangle (1.1,0.44);
      \draw[dashline] (0,0) -- (3.6,0);
    \end{scope}

    \draw[dashline] (rear) -- ($(front)+(2.4,0)$);
    \draw[line width=1pt] (rear) -- (front);

    \draw[force] (com) -- ++(1.705,1.15) node[above right=-2pt, stdstate]{$v$};
    \draw ($(com)+(0.85,0)$) arc (0:\slip:0.85);
    \node[stdstate] at (\slip*0.5:1.1) {$\beta$};

    \fill (com) circle (3pt);
    \draw[-{Latex[length=2.6mm]}, thick]
        ($(com)+(-0.5,0)$) arc (-180:140:0.5);
    \node[stdstate] at (90:0.8) {$\dot{\Psi}$};

    \draw[force] ($(rear)+(-\fxlen,0)$) -- (rear);
    \node[left] at ($(rear)+(-1.7,0)$) {$F_{xR}$};
    \draw[force] ($(rear)+(0,-\fylen)$) -- (rear);
    \node[below] at ($(rear)+(0,-\fylen-0.1)$) {$F_{yR}$};
    \pic[transform shape] at ($(rear)+(0,\spinat)$) {wheelspin};
    \node[right, stdstate] at ($(rear)+(\spinrx,\spinat)$) {$\omega_r$};

    \begin{scope}[shift={(front)}, rotate=\stang]
      \draw[force] (-\fxlen,0) -- (0,0);
      \node[anchor=north west] at (-\fxlen,-0.05) {$F_{xF}$};
      \draw[force] (0,-\fylen) -- (0,0);
      \node[anchor=north west] at (0.05,-\fylen) {$F_{yF}$};
      \pic[transform shape] at (0,\spinat) {wheelspin};
      \node[right, stdstate] at (\spinrx,\spinat) {$\omega_f$};
    \end{scope}

    \draw[dashline] (rear) -- ($(rear)+(2.7,0.93)$);
    \draw ($(rear)+(1.8,0)$) arc (0:19:1.8);
    \node at ($(rear)+(2.15,0.38)$) {$\alpha_R$};

    \begin{scope}[shift={(front)}]
      \draw[dashline] (0,0) -- (\stang+16:2.9);
      \draw (\stang:2.0) arc (\stang:\stang+16:2.0);
      \node at (\stang+8:2.4) {$\alpha_F$};
      \draw (2.2,0) arc (0:\stang:2.2);
      \node[stdstate] at (\stang*0.5:2.5) {$\delta$};
    \end{scope}
  \end{scope}

  \draw (0.9,0) arc (0:\yaw:0.9);
  \node[stdstate] at (10:1.15) {$\Psi$};

  \begin{scope}[thesisred]
    \coordinate (foot) at ({\ey*sin(\pathang)}, {-\ey*cos(\pathang)});
    \coordinate (pathc) at ($(foot)+({-\pathR*sin(\pathang)}, {\pathR*cos(\pathang)})$);

    \draw[line width=1pt, dashed]
        ($(pathc)+({\pathang-90-3}:\pathR)$) arc ({\pathang-90-3}:{\pathang-90+26}:\pathR)
        node[right, align=left, font=\small] {reference\\[-2pt]path};

    \draw[-{Latex[length=2mm]}, thick] (foot) -- node[left=1pt] {$e_y$} (0,0);
    \draw ($(foot)!0.25cm!(0,0)$) -- ++(\pathang:0.25cm) -- ++({\pathang-90}:0.25cm);

    \draw[dashline] (0,0) -- (\pathang:3.4)
        node[right, align=left, font=\small] {path\\[-2pt]tangent};

    \draw[-{Latex[length=2mm]}, thick] (\pathang:1.6) arc (\pathang:\yaw:1.6);
    \node at (-19:1.36) {$e_\Psi$};
  \end{scope}
\end{tikzpicture}}
  \caption[Single-track bicycle model]{The single-track model. The STD states (blue) include the Cartesian position $(s_x, s_y)$ of the CG in the global frame; the Frenet-frame pose relative to a reference path (red) is given by the cross-track error $e_y$ and the heading error $e_\Psi$. Variables are defined in Table~\ref{tab:bicycle_vars}.}
  \label{fig:bicycle_model}
\end{figure}

A tire generates lateral force as a function of its slip angle $\alpha$. Slip angle is the misalignment between the tire's direction of travel and its orientation, shown as $\alpha_F$ and $\alpha_R$ for the front and rear axles in Fig.~\ref{fig:bicycle_model}. Notice that the velocities at the front and rear tires are not parallel to the velocity $v$ at the centre of gravity (CG) because the vehicle is also rotating: the yaw rate $\dot\Psi$ adds a lateral component at each axle that grows with its distance from the CG and has opposite signs at the two axles. Fig.~\ref{fig:bicycle_model} shows counter-clockwise rotation. 

The relationship between lateral force and slip angle has three regions: 1) a linear rise at small slip angles, whose slope is the \emph{cornering stiffness}; 2) a peak near the friction limit; and 3) a decay beyond it as the tire begins to slide. This last region is termed \emph{tire saturation}, and it is where a drifting vehicle operates. The relationship is commonly captured by the Pacejka tire model~\cite{pacejkaMagicFormulaTyre1992, TireVehicleDynamics2012}. Similarly, longitudinal force is generated by longitudinal slip, the mismatch between how fast the wheel spins and how fast it travels. 

The lateral and longitudinal forces are not independent: their resultant is bounded by the friction circle of radius $\mu F_z$, for $\mu$ the friction coefficient and $F_z$ the normal load on the tire~\cite{hindiyehDrifting}, so longitudinal force spent at a tire reduces the lateral force it can still generate, and vice versa. Tire models that resolve this coupling are termed combined-slip; the lateral and longitudinal forces at the rear axle are shown as $F_{yR}$ and $F_{xR}$ in Fig.~\ref{fig:bicycle_model}.

A drifting state is characterized by a large sideslip angle $\beta$, the angle between the velocity vector at the CG and the chassis centreline, and a large yaw rate $\dot\Psi$, the rate of rotation about the vertical axis; it typically involves driven tires held in saturation, and counter-steer. A drift is an equilibrium of the vehicle dynamics~\cite{milaniVehicleDriftingDynamics2022}, held by throttle and steering acting on the yaw moment in opposite directions. Throttle increases the longitudinal force at the driven tires, shown as $F_{xR}$ at the rear axle in Fig.~\ref{fig:bicycle_model}, which reduces the lateral force those tires can still generate and so increases the yaw moment; counter-steer decreases the front slip angle $\alpha_F$, reducing the lateral force $F_{yF}$ ahead of the CG and with it the yaw moment.

Expressing this maneuver requires a model that resolves the combined-slip tire forces described above. We select a single-track chassis model, in which the front and rear axle tires are each lumped into one, with a combined-slip Pacejka PAC2002 tire model~\cite{AdamsTireUsers2011, TireVehicleDynamics2012}, following the implementation in the commonroad-vehicle-models
module~\cite{althoffCommonRoadComposableBenchmarks2017}, where it is termed the single-track drift (STD) model. Similar models are used throughout the autonomous racing and drifting literature~\cite{liuAutonomousDriftingProfessional2025}.

The STD model carries a nine-element state vector,
\begin{equation}
    \label{eq:std_state}
    \mathbf{x} = \big[\, s_x,\; s_y,\; \delta,\; v,\; \Psi,\; \dot{\Psi},\; \beta,\; \omega_f,\; \omega_r \,\big]^{\top},
\end{equation}
whose elements are defined in Table~\ref{tab:bicycle_vars} and shown graphically in Fig.~\ref{fig:bicycle_model} in blue. The last two, the front and rear wheel angular velocities $\omega_f$ and $\omega_r$, are what make the model drift-capable: they give the longitudinal slip at each tire, which the PAC2002 formula needs to resolve the combined lateral and longitudinal forces bounded by the friction circle. The model is driven by two inputs, the steering rate $\dot\delta$ and the longitudinal acceleration $a$, also listed in Table~\ref{tab:bicycle_vars}. For the full equations of motion, tire formulas, and 53 parameters, we refer the reader to the CommonRoad benchmark~\cite{althoffCommonRoadComposableBenchmarks2017}.

STD states place the vehicle in a global Cartesian frame. For path tracking it is also convenient to express the pose in the Frenet frame, which measures it relative to the reference path rather than to a global origin, and two Frenet quantities are used throughout this paper: the cross-track error $e_y$, the signed perpendicular distance from the vehicle to the path, and the heading error $e_\Psi$, the angle between the vehicle heading $\Psi$ and the path tangent. Both follow from the Cartesian pose $(s_x, s_y, \Psi)$ given a reference path.

\begin{table}[t]
    \centering
    \footnotesize
    \renewcommand{\arraystretch}{1.15}
    \caption{Single-Track Model Quantities}
    \label{tab:bicycle_vars}
    \begin{tabular}{|c|c|c|}
        \hline
        \textbf{Quantity} & \textbf{Symbol} & \textbf{Units} \\
        \hline
        \multicolumn{3}{|l|}{\textit{Variables}} \\
        \hline
        Cartesian coordinates of CG     & $s_x,\ s_y$ & m    \\
        Front axle steering angle       & $\delta$   & rad   \\
        Velocity at CG                  & $v$        & m/s   \\
        Vehicle heading                 & $\Psi$     & rad   \\
        Yaw rate                        & $\dot\Psi$ & rad/s \\
        Body sideslip angle             & $\beta$    & rad   \\
        Front wheel angular velocity    & $\omega_f$ & rad/s \\
        Rear wheel angular velocity     & $\omega_r$ & rad/s \\
        \hline
        \multicolumn{3}{|l|}{\textit{Derived variables}} \\
        \hline
        Front axle slip angle         & $\alpha_F$ & rad   \\
        Rear axle slip angle            & $\alpha_R$ & rad   \\
        Front axle lateral force        & $F_{yF}$   & N     \\
        Rear axle lateral force         & $F_{yR}$   & N     \\
        Rear axle longitudinal force    & $F_{xR}$   & N     \\
        Cross-track error               & $e_y$      & m     \\
        Heading error                   & $e_\Psi$   & rad   \\
        \hline
        \multicolumn{3}{|l|}{\textit{Inputs}} \\
        \hline
        Steering rate                   & $\dot\delta$ & rad/s   \\
        Longitudinal acceleration       & $a$          & m/s$^2$ \\
        \hline
    \end{tabular}
\end{table}

\subsection{The Recovery Task}
\label{sec:recovery_task}
\begin{figure}[t]
  \centering
  \includegraphics{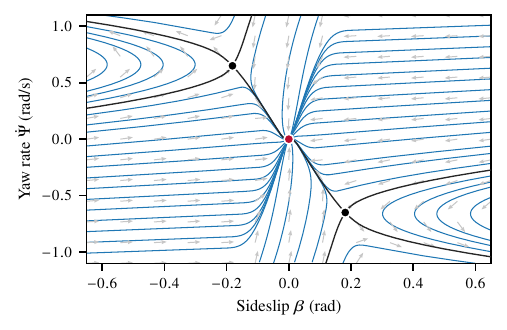}
  \caption[Sideslip vs. yaw-rate phase portrait]{Sideslip vs. yaw-rate phase plane of
  Hindiyeh's two-state single-track model~\cite{hindiyehDrifting}.}
  \label{fig:phase-plane}
\end{figure}

Recovery from an unstable vehicle state requires guiding the vehicle away from the unstable regime and towards a stable equilibrium corresponding to conventional driving. Equilibria of vehicle dynamics such as those studied in this paper can be identified using the $\beta$--$\dot\Psi$ phase plane of Fig.~\ref{fig:phase-plane}. The plane in the figure was built on Hindiyeh's two-state single-track model, which has been used to design a drift controller for a full-scale vehicle~\cite{hindiyehDrifting}. The red dot at the origin is the stable equilibrium, with no sideslip and no yaw rate, corresponding to driving straight along the path. The black dots are the drift equilibria, each a saddle point~\cite{hindiyehDrifting, bobier-tiuVehicleControlSynthesis2019}, which is why both throttle and steering must be applied actively for a drift to be held. Arrows denote the direction in which the state evolves. From many states it moves away from every equilibrium, the stable one included, which motivates active control to bring the vehicle back towards it.

Prior work on learned policies that tackled the problem has used relatively loose or empirical definitions of recovery, as discussed in Section~\ref{sec:related_safety}. We use phase plane analysis to define recovery with respect to the stable equilibrium associated with conventional driving. The phase plane therefore provides the physical basis for the recovery criterion, rather than defining recovery solely through an empirically selected duration or terminal condition.

Because reaching the stable equilibrium in the $(\beta,\dot\Psi)$ phase plane alone does not guarantee that the vehicle is in a practically recovered state, we augment this equilibrium condition with bounds on the steering input, state derivatives, and tracking errors. Specifically, we impose five conditions on the vehicle's dynamic state and two additional conditions on its alignment with the reference path.
  
Let $\mathbf{z}_t = (\delta,\, \beta,\, \dot\Psi,\, \dot\beta,\, \ddot\Psi,\, e_\Psi,\, e_y)$ denote the vector of recovery-relevant quantities, assembled from the STD state $\mathbf{x}_t$ of \eqref{eq:std_state}, its time derivatives, and the vehicle's Frenet-frame pose relative to the reference path. Here we adopt the STD-model notation $\dot\Psi$ for the yaw rate and $\ddot\Psi$ for the yaw acceleration, respectively $r$ and $\dot r$ in the phase-plane literature. A vehicle state is recovered if and only if $\mathbf{z}_t \in \mathcal{R}$, given by

\begin{equation}
    \label{eq:recovery_set}
    \mathcal{R} = \left\{\, \mathbf{z} \;\middle|\;
    \begin{aligned}
        &|\delta| \le \delta_{\mathrm{th}}, \;\; |\beta| \le \beta_{\mathrm{th}}, \\
        &|\dot\Psi| \le \dot\Psi_{\mathrm{th}}, \;\; |\dot\beta| \le \dot\beta_{\mathrm{th}}, \\
        &|\ddot\Psi| \le \ddot\Psi_{\mathrm{th}}, \;\; |e_\Psi| \le e_{\Psi,\mathrm{th}}, \\
        &|e_y| \le e_{y,\mathrm{th}}
    \end{aligned}
    \,\right\}.
\end{equation}

The first five conditions characterize the vehicle's dynamic state around the stable operating condition, while the last two ensure that the vehicle is aligned with and remains sufficiently close to the reference path. The bounds on $\dot\beta$ and $\ddot\Psi$ in particular require the stable state to be held rather than crossed transiently, which is why reaching $\mathcal{R}$ once suffices. Each $(\cdot)_{\mathrm{th}}$ denotes a threshold, a design choice set by the vehicle's operating conditions and the tolerance for considering it recovered; the values we use are given in Section~\ref{sec:baselines_metrics}.

By extension, the set $\mathcal{U} = \mathcal{R}^c$ is the set of non-recovered states from which recovery episodes begin. $\mathcal{U}$ is unbounded and partly unrecoverable, and the boundary of its recoverable part is not known in closed form.

We formalize the \textbf{recovery task} as: given a vehicle travelling along a straight reference path in an uncontrolled initial state $\mathbf{z}_0 \in \mathcal{U}$, find a sequence of control commands that drives the vehicle to a recovered state $\mathbf{z}_T \in \mathcal{R}$, where $T$ is the first time at which $\mathcal{R}$ is reached, subject to $|e_y| \le e_{y,\mathrm{th}}$ for all $t$. Thus, a recovery episode is successful when the vehicle reaches $\mathcal{R}$ without violating the lateral tracking constraint during the recovery process. The primary objective is to maximize the measure of initial conditions in $\mathcal{U}$ from which the vehicle can be successfully recovered, and the secondary objective is to minimize recovery time $T$.

\section{Recovery Controller Design}
\label{sec:method}

\begin{figure*}[!tb]
  \centering
  \input{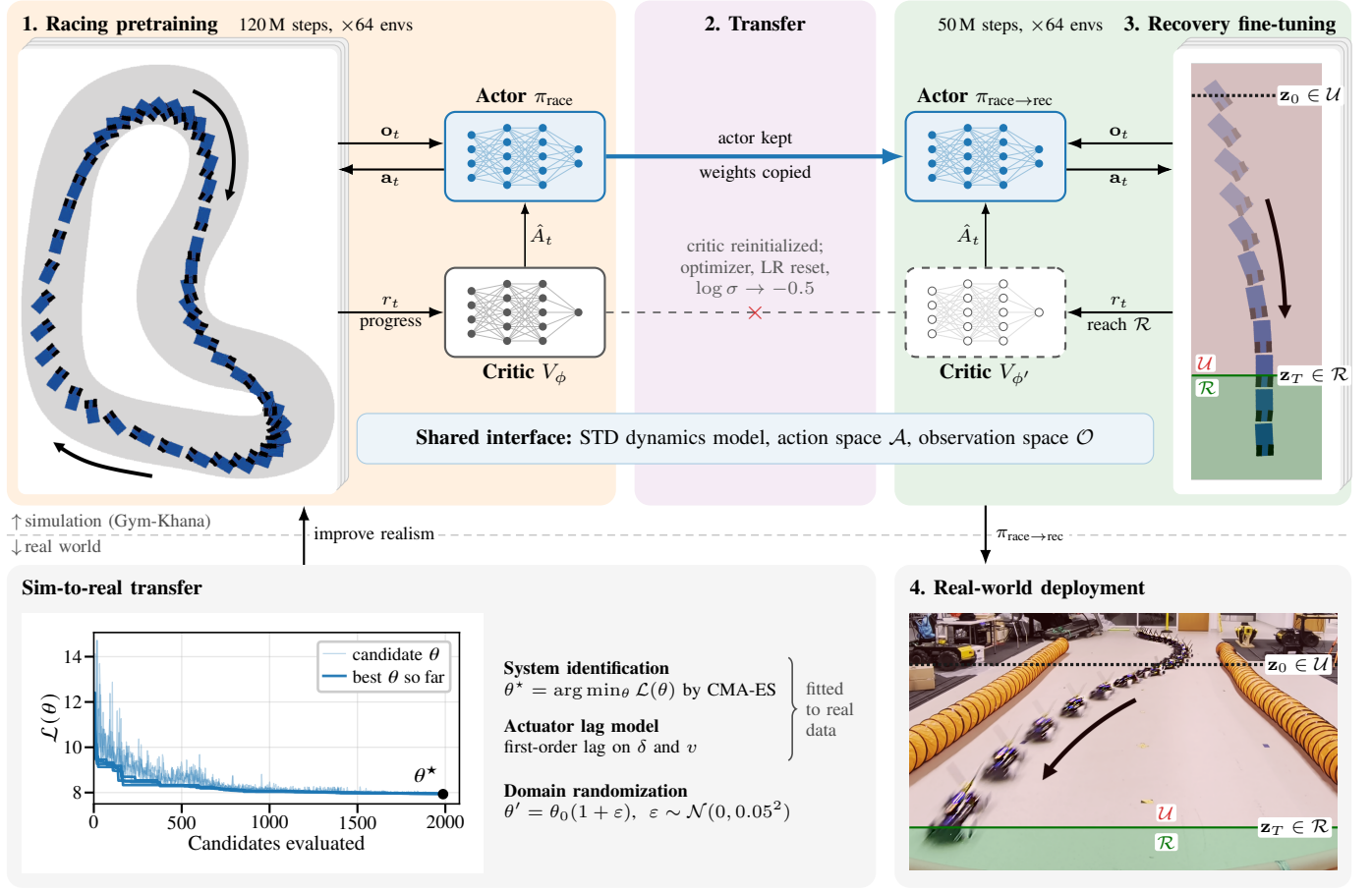}
  \caption{Overview of the approach. \textbf{1.}~A policy is trained with proximal policy optimization (PPO) to race on the Drift track and its mirror image, where drifting emerges as the time-optimal technique. \textbf{2.}~Its actor $\pi_{\text{race}}$ is transferred unchanged to the recovery task, and its critic is reinitialized. \textbf{3.}~Training continues under the recovery reward, which drives the vehicle from an unrecovered state in $\mathcal{U}$ into the recovered set $\mathcal{R}$; the two tasks share one vehicle model, observation and action space, so only the reward changes between them. \textbf{4.}~The resulting $\pi_{\text{race}\to\text{rec}}$ is deployed on the physical F1TENTH vehicle. Before deployment, system identification, actuator-lag modelling and domain randomization adapt the simulator to the vehicle, and the controller is retrained on it from racing onwards (Sections~\ref{sec:sim2real-transfer} and~\ref{sec:hardware_validation}). Every image is taken from our experiments: the racing and recovery panels are rollouts of the trained policies in simulation, the bottom-left plot is the loss \eqref{eq:sysid_objective} of the identification run across four parallel covariance matrix adaptation evolution strategy (CMA-ES) workers, and the bottom-right photo overlays frames of a recovery run on the physical vehicle. In both recovery panels time runs downwards: the shading marks $\mathcal{U}$ (red) and its complement $\mathcal{R}$ (green), the dotted line the initial state $\mathbf{z}_0$ where the policy takes over, and the solid line the first recovered state $\mathbf{z}_T$.}
  \label{fig:approach}
\end{figure*}

The nonlinear dynamics of the related racing and recovery tasks are well-suited to a learned RL policy. Fig.~\ref{fig:approach} gives an overview of our approach. We first define an RL interface that is applicable to both tasks, consisting of a shared vehicle model, observation vector, and action output. We then train a policy to race in simulation, where it discovers drifting as the time-optimal control technique, and explores a broad region of vehicle behaviours, including how to initiate, sustain, and exit a drift. This pretraining encodes vehicle dynamics information into the policy, which is then transferred to the recovery task. Both the racing and recovery policies are trained in a purpose-built simulator, Gym-Khana, which builds on the lightweight f1tenth\_gym~\cite{okellyF1TENTHOpensourceEvaluation2020} and implements the vehicle model and track in a Gymnasium-compliant, parallelized environment. Finally, we validate the controller on a physical F1TENTH vehicle; the steps that adapt the simulator to it are given in Section~\ref{sec:sim2real-transfer}, and the hardware results in Section~\ref{sec:hardware_validation}.

\subsection{Observations and Actions}
\label{sec:obs_act}
In order to train a policy for the racing task, and then transfer it to the recovery task, it is necessary that the core components are shared. In addition to the vehicle parameters and dynamics model, this includes the RL observation space and action space. The action space, the output of the actor network, is chosen as a two-dimensional vector:
\begin{equation}
    \mathbf{a}_t = [\delta_{\text{ref}}, v_{\text{ref}}] \in \mathcal{A},
    \label{eq:racing_action}
\end{equation}
representing the continuous action space of commanded steering angle and velocity. Low-level controllers map these to the STD inputs $\dot\delta$ and $a$ in the Gym-Khana simulator.

The observation vector provides the policy with the vehicle and track state information required to race. It combines vehicle state, expressed in the notation of the STD model, with a local description of the upcoming track:
\begin{equation}
    \mathbf{o}_t = \big[\, v_x,\; v_y,\; e_\Psi,\; e_y,\; \dot\Psi,\; \beta,\; \omega_{\text{avg}},\; \boldsymbol{\kappa},\; \mathbf{w} \,\big] \in \mathcal{O}.
    \label{eq:racing_obs}
\end{equation}

The yaw rate $\dot\Psi$, sideslip $\beta$ and Frenet errors $e_\Psi$, $e_y$ are defined in Section~\ref{sec:model}. The velocities $v_x$ and $v_y$ are the longitudinal and lateral components of $v$ in the vehicle frame, and the wheel speed $\omega_{\text{avg}}$ is the average of $\omega_f$ and $\omega_r$ in \eqref{eq:std_state}. 
The upcoming track geometry is represented by $N$ centreline points sampled uniformly ahead of the vehicle, with $N=5$ and a spacing of $0.5$~m. The curvature $\boldsymbol{\kappa} \in \mathbb{R}^N$ is computed at each sampled point, providing a representation of how the track bends along the upcoming path. In contrast, the track width $\mathbf{w} \in \mathbb{R}^2$ is measured only at the first and last sampled points, capturing the available track width over the lookahead region without unnecessarily increasing the observation dimension.
The full observation space, therefore, lies in $\mathcal{O} \subseteq \mathbb{R}^{9 + N}$. The entire observation space is designed to be agnostic of map-frame track location, enabling the policy to learn a generalizable racing behaviour. Both action and observation spaces are normalized to $[-1,1]$, as is best practice for deep RL algorithms~\cite{andrychowiczWhatMattersOnpolicy2021}. 

\subsection{Racing Pretraining}
\label{sec:racing_policy}

We define the racing task as maximizing progress along a reference centreline without leaving the track, rewarding per-step progress in Frenet arc-length with a penalty for leaving the track boundaries~\cite{czechmanowskiLearningRacingPolicies2025b, fuchsSuperHumanPerformanceGran2021}. No raceline or driving style is prescribed, so the policy discovers them through trial and error, including drifting, the time-optimal technique through a corner on a low-friction surface~\cite{taverniniMinimumTimeCornering2013} (Fig.~\ref{fig:approach}, panel~1).

To match the F1TENTH testbed, we take chassis parameters from the f1tenth\_gym project~\cite{okellyF1TENTHOpensourceEvaluation2020} and tire parameters from the Adams user handbook~\cite{AdamsTireUsers2011}. In the simulation environment we create a map with a mix of corners and straights, adapted from the work of Czechmanowski et al., called the ``Drift'' track~\cite{czechmanowskiLearningRacingPolicies2025b}. To learn a policy symmetric for left and right cornering actions, we train equally on the original track and a mirrored copy, in both directions.

We formulate the task episodically and distinguish two ways an episode can end. It \emph{terminates} when the vehicle leaves the track, a true end of the task, so the value target at the final step is the reward alone. It is \emph{truncated} after 4096 timesteps, a limit that only bounds the training rollout and is not part of the task, so the value target instead bootstraps from the critic's estimate of the final state. Per-step progress is a surrogate for lap time: maximizing it is equivalent to maximizing speed along the reference centreline. On termination or truncation an episode is reset by placing the agent at rest at a position along the centreline that is uniformly sampled.

\begin{table}[t]
    \centering
    \footnotesize
    \renewcommand{\arraystretch}{1.15}
    \caption{PPO Hyperparameters Modified from SB3 Defaults}
    \label{tab:ppo_hyperparams}
    \begin{tabular}{|c|c|}
        \hline
        \textbf{Hyperparameter} & \textbf{Value} \\
        \hline
        Parallel environments       & 64 \\
        Total training timesteps    & $120 \times 10^{6}$ \\
        Rollout length              & 4096 timesteps \\
        Timestep                    & $0.01$~s ($100$~Hz) \\
        Learning rate (linear)      & $3 \times 10^{-4} \rightarrow 1 \times 10^{-4}$ \\
        Actor network               & MLP, two hidden layers of 128, tanh \\
        Critic network              & MLP, two hidden layers of 256, tanh \\
        Log-std schedule (linear)   & $-1.0 \rightarrow -3.0$ \\
        \hline
    \end{tabular}
\end{table}

We train the policy with proximal policy optimization (PPO)~\cite{schulmanProximalPolicyOptimization2017}, implemented in Stable-Baselines3 (SB3)~\cite{raffinStableBaselines3ReliableReinforcement}. PPO is used for its on-policy formulation and robustness to hyperparameter choice. We use the SB3 defaults for all hyperparameters except those listed in Table~\ref{tab:ppo_hyperparams}, including the multi-layer perceptron (MLP) sizes.

\subsection{Racing-to-Recovery Transfer}
\label{sec:transfer}

Our recovery controller is the racing policy $\pi_{\text{race}}$ of Section~\ref{sec:racing_policy} transferred to the recovery task of Section~\ref{sec:recovery_task}, which we denote $\pi_{\text{race}\to\text{rec}}$ (Fig.~\ref{fig:approach}, panels~2--3). With the interface of Section~\ref{sec:obs_act} shared, the two tasks differ only in their reward and their episodic setup. 

Each episode places the vehicle on the centreline of a straight lane with the four perturbed quantities $(v_0,\, \beta_0,\, \dot\Psi_0,\, e_{\Psi,0})$ sampled uniformly from the ranges of Table~\ref{tab:recovery_init_ranges} and embedded into the STD state \eqref{eq:std_state}. It terminates on entering the recovered set $\mathcal{R}$ \eqref{eq:recovery_set} and on lane departure, $|e_y| > e_{y,\mathrm{th}}$, which enforces the lateral constraint continuously rather than only on arrival at $\mathcal{R}$; both are true terminal states, so no value bootstrapping is applied at either. The threshold values defining $\mathcal{R}$ are listed in Table~\ref{tab:recovery_thresholds}. Episodes are truncated, with bootstrapping, after 2048 timesteps or 46~m of travel along the path.

The reward has three components serving the task's two objectives: 1) recovering from as many initial states as possible, and 2) recovering in minimal time. The recovery reward $r^\text{rec}_t$ and the out-of-bounds penalty $r^\text{oob}_t$ serve the first objective, and the constant per-step penalty $r^\text{step}_t$ serves the second:
\begin{equation}
    \begin{aligned}
        r_t &= r^\text{step}_t + r^\text{rec}_t + r^\text{oob}_t, \\[2pt]
        r^\text{step}_t &= -K_\text{step}, \\[2pt]
        r^\text{rec}_t &=
        \begin{cases}
            r_\text{rec}, & \text{if } \mathbf{z}_t \in \mathcal{R}, \\
            0, & \text{otherwise,}
        \end{cases} \\[2pt]
        r^\text{oob}_t &=
        \begin{cases}
            r_\text{oob}, & \text{if } |e_y| > e_{y,\mathrm{th}}, \\
            0, & \text{otherwise,}
        \end{cases}
    \end{aligned}
    \label{eq:recovery_reward}
\end{equation}
with $K_\text{step} = 0.01$, $r_\text{rec} = 100$, and $r_\text{oob} = -50$.

The transfer retains the learned control representation and discards what was fitted to the racing reward. The actor is kept unchanged: its mapping from vehicle and track state to steering and velocity commands is the knowledge to be reused. Four components are reset. 1) The critic is reinitialized with orthogonal weights, since a value function fit to racing returns predicts the wrong quantity under the recovery reward. 2) The Adam optimizer state is cleared of moments accumulated on the racing objective. 3) The learning rate is restarted on the same linear decay used for racing, the source run having already annealed to its minimum. 4) The log-standard-deviation $\log\sigma$ of the policy's Gaussian action noise is reset from its converged value to $\log\sigma = -0.5$, wider than the $-1.0$ racing began from, so the near-deterministic racing policy is pushed to discover recovery behaviour rather than replay racing behaviour. Training then continues under the recovery reward for a further 50 million timesteps.

\subsection{Sim-to-Real Transfer}
\label{sec:sim2real-transfer}
To enable validation of the trained policies in the real world, a procedure for fine-tuning the parameters in the simulated environment is needed.
The STD model parameters are matched to the real testbed in two ways, according to how each can be measured. Those that are directly measurable are measured: the vehicle mass $m$, the wheelbase measurements, the width and length dimensions, the front/rear torque split, and the mechanically limited steering angles. All remaining chassis, actuator lag and tire parameters cannot be directly measured. 
We collect real data from the vehicle under manual control, replay the same control inputs in simulation, and select the parameters $\theta$ that minimize the difference between the two. The vector $\theta$ holds 29 parameters: the yaw and wheel moments of inertia, the centre-of-gravity height, the effective wheel radius, the maximum acceleration and steering rate, the steering and speed lag time constants $T_\mathrm{steer}$ and $T_\mathrm{speed}$ of \eqref{eq:first_order_lag}, and 21 PAC2002 tire coefficients. The objective is
\begin{equation}
    \begin{aligned}
        \theta^\star &= \arg\min_{\theta} \; \mathcal{L}(\theta), \\
        \mathcal{L}(\theta) &= \sum_{c \in \mathcal{C}} k_c \, \big\| \mathbf{x}^{\text{sim}}_c(\theta) - \mathbf{x}^{\text{real}}_c \big\|^2,
    \end{aligned}
    \label{eq:sysid_objective}
\end{equation}
a weighted mean-squared error between the simulated and real signals over a set of channels $\mathcal{C}$: the body velocities $v_x, v_y$, the yaw rate $\dot\Psi$, the average wheel speed $\omega_{\text{avg}}$, the Cartesian pose $(s_x, s_y, \Psi)$, and the sideslip $\beta$, with weights $k_c$ that approximately normalize their differing magnitudes. Logs are resampled to the 100~Hz control rate and split into windows of 1.5~s, each reset to the measured state and then applied in open loop, so that the fit targets the short-horizon accuracy on which a closed-loop controller depends. Because the simulator is a non-differentiable numerical integrator, we treat $\mathcal{L}$ as a black box and minimize it with the covariance matrix adaptation evolution strategy (CMA-ES)~\cite{hansenCompletelyDerandomizedSelfadaptation2001}, evaluating runs in parallel with Optuna~\cite{akibaOptunaNextgenerationHyperparameter2019}; the bottom-left plot of Fig.~\ref{fig:approach} shows the loss of one such run across four parallel workers.
The identified parameters are validated by comparing simulated and measured trajectories on held-out data not used in the optimization.

We model the actuator lag for both steering and velocity actions, as this has been demonstrated to aid in sim-to-real transfer~\cite{czechmanowskiLearningRacingPolicies2025b}. Both follow a first-order lag structure, relating the reference inputs $\delta_\mathrm{ref}, v_\mathrm{ref}$ of \eqref{eq:racing_action} to the actual values $\delta, v$:
\begin{equation}
    \begin{aligned}
        \dot\delta &= \frac{1}{T_\mathrm{steer}} \left( \delta_\mathrm{ref} - \delta \right), \\
        \dot v &= \frac{1}{T_\mathrm{speed}} \left( v_\mathrm{ref} - v \right),
    \end{aligned}
    \label{eq:first_order_lag}
\end{equation}
where the time constants $T_\mathrm{steer}$ and $T_\mathrm{speed}$ capture the real vehicle's lag and are identified as part of $\theta$.

Identification narrows the sim-to-real gap but does not close it, as model accuracy is limited and the real dynamics change with battery level and tire wear. Previous work has underscored the importance of applying minimal, targeted domain randomization to enable the policy to generalize without overly degrading performance~\cite{czechmanowskiLearningRacingPolicies2025b, zhaoSimtoRealTransferDeep2020}. We therefore randomize only where it is warranted. The actuator parameters, the steering lag constant and the steering-rate and acceleration limits, are randomized because the lack of steering feedback leaves them the least certain part of the model; the four tire parameters with the greatest influence on the dynamics are randomized because the controller must be robust under tire-saturated drifting. At the start of each training episode, each randomized parameter with nominal value $\theta_0$ is perturbed and held fixed for the episode:
\begin{equation}
    \theta' = \theta_0\,(1 + \varepsilon), \qquad
    \varepsilon \sim \mathcal{N}(0, \sigma^2), \quad |\varepsilon| \le 3\sigma,
    \label{eq:domain_rand}
\end{equation}
with $\sigma = 0.05$, a $\pm 5\%$ perturbation clipped at $\pm 15\%$, drawn each time from the fixed nominal value rather than compounded across episodes. Randomization is applied during training only; evaluation uses the nominal parameters, to keep metrics comparable across runs.

The combination of these procedures enables us to validate the superior performance of our proposed approach in the real world, as presented in Section~\ref{sec:hardware_validation}.

\section{Simulation Experiments}
\label{sec:sim_results}
This section evaluates the recovery controllers in simulation, first establishing the evaluation metrics and experimental protocol, and then comparing the proposed controller against classical baselines and alternative training strategies.

\subsection{Baselines and Metrics}
\label{sec:baselines_metrics}

We choose two classical control methods as baselines: a geometric path-tracking Stanley controller and a nonlinear MPC formulation.
The former, Stanley~\cite{thrunStanleyRobotThat2006}, is a controller that uses a weighted sum of the heading error $e_\Psi$ and the cross-track error $e_y$, the two terms that suit it to the recovery task. In our work, it tracks the straight reference path, with its gains tuned experimentally for that task. It is steering-only, so for longitudinal control we command the constant velocity $v_0$ of the initial state, which is the controller's primary limitation: it cannot modulate braking or acceleration as part of the recovery strategy.

The second baseline is a Frenet-frame MPC based on the formulation used in the ForzaETH race stack~\cite{baumannForzaETHRaceStack2025}. The MPC problem is formulated using CasADi and solved online with the acados optimal-control solver~\cite{verschuerenAcadosModularOpensource2021, anderssonCasADiSoftwareFramework2019}. It uses a single-track vehicle model with a lateral-force-only Pacejka tire model, and we therefore refer to it as the single-track MPC (ST-MPC).
Its cost penalizes lateral deviation, heading error, velocity-tracking error and lateral velocity, and rewards path progress, subject to boundary and friction-ellipse constraints; penalizing lateral velocity, and so sideslip, is what suits it to recovery. Its limitation is the one common to real-time MPC: it cannot afford the full combined-slip PAC2002 tire model our learned policies are trained on due to the model complexity. 

Both baselines reflect conservative handling-regime assumptions of production safety controllers, and we show that the learned controller outperforms both. We do not claim to beat the best drift-capable MPC: the performance of RL against state-of-the-art MPC at the limit of handling is a question prior work has already addressed (Section~\ref{sec:related_limit}).

In defining the recovery task in simulation, we must first choose two sets of values: the initial-state sampling ranges that define the breadth of $\mathbf{z}_0 \in \mathcal{U}$ during training (Section~\ref{sec:transfer}), and the threshold values that define the recovered set $\mathcal{R}$ \eqref{eq:recovery_set} (Section~\ref{sec:recovery_task}). These are given in Tables~\ref{tab:recovery_init_ranges} and~\ref{tab:recovery_thresholds}, respectively.

\begin{table}[t]
    \centering
    \footnotesize
    \renewcommand{\arraystretch}{1.15}
    \setlength{\tabcolsep}{4pt}
    \caption{Initial-State Sampling Ranges for Training}
    \label{tab:recovery_init_ranges}
    \begin{tabular}{|c|c|c|c|}
        \hline
        \textbf{Quantity} & \textbf{Symbol} & \textbf{Range} & \textbf{Equivalent} \\
        \hline
        Velocity          & $v_0$          & $[2,\ 12]$~m/s           & --- \\
        Sideslip angle    & $\beta_0$      & $[-0.349,\ 0.349]$~rad   & $\pm 20^\circ$ \\
        Yaw rate          & $\dot\Psi_0$   & $[-0.785,\ 0.785]$~rad/s & $\pm 45^\circ$/s \\
        Heading error     & $e_{\Psi,0}$   & $[-0.785,\ 0.785]$~rad   & $\pm 45^\circ$ \\
        \hline
    \end{tabular}
\end{table}

\begin{table}[t]
    \centering
    \footnotesize
    \renewcommand{\arraystretch}{1.15}
    \setlength{\tabcolsep}{4pt}
    \caption{Recovered-Set Thresholds}
    \label{tab:recovery_thresholds}
    \begin{tabular}{|c|c|c|c|}
        \hline
        \textbf{Quantity} & \textbf{Symbol} & \textbf{Value} & \textbf{Equivalent} \\
        \hline
        Steering angle    & $\delta_{\mathrm{th}}$     & $0.175$~rad      & ${\approx}\,10^\circ$ \\
        Sideslip angle    & $\beta_{\mathrm{th}}$      & $0.07$~rad       & ${\approx}\,4^\circ$ \\
        Yaw rate          & $\dot\Psi_{\mathrm{th}}$   & $0.175$~rad/s    & ${\approx}\,10^\circ$/s \\
        Sideslip rate     & $\dot\beta_{\mathrm{th}}$  & $7.0$~rad/s      & --- \\
        Yaw acceleration  & $\ddot\Psi_{\mathrm{th}}$  & $17.5$~rad/s$^2$ & --- \\
        Heading error     & $e_{\Psi,\mathrm{th}}$     & $0.087$~rad      & ${\approx}\,5^\circ$ \\
        Cross-track error & $e_{y,\mathrm{th}}$        & $1.1$~m          & $1/2$ lane width \\
        \hline
    \end{tabular}

    \vspace{2pt}
    \parbox{\columnwidth}{\scriptsize The two rate thresholds are each the
    corresponding state threshold divided by the $0.01$~s timestep,
    admitting a state that would traverse half its tolerance band in one timestep.}
\end{table}

Next, we define the recovery metrics that enable a comparison of the various approaches. As no formal guarantee is available for a learned controller, we compare experimentally, running every controller on a common set of representative initial states in $\mathcal{U}$. We evaluate all controllers on a total of $N_\text{tot} = 294$ initial states defined by $(v_0,\, \beta_0,\, \dot\Psi_0,\, e_{\Psi,0})$, as described in Section~\ref{sec:transfer}. The grid ranges are given in Table~\ref{tab:recovery_eval_grid}: the values of each of the four variables are spaced uniformly within its range, and every combination is evaluated, giving $N_\text{tot} = 7 \times 7 \times 2 \times 3 = 294$.

The evaluation grid deliberately spans a far wider range of initial conditions than the training distribution of Table~\ref{tab:recovery_init_ranges}, as we wish to evaluate the entire breadth of recoverable states. For our learned policies, this involves generalizing to loss-of-control states more extreme than any encountered while learning. The four state bounds were set by choosing representative velocity and heading-error ranges, and then expanding the sideslip and yaw-rate ranges gradually, symmetric about the origin, until the corners of the grid, the most extreme initial conditions, were $0\%$ recoverable for all controllers, thus spanning all reasonable initial states. The $(\beta_0, \dot\Psi_0)$ grid forms the axes of the heatmaps in Fig.~\ref{fig:recovery_sim_comparison}.

Evaluation follows the recovery task protocol described in Section~\ref{sec:transfer}. We analyze two properties across the grid: the recovery rate, measuring efficacy, and the recovery time, measuring efficiency. These address the two objectives of the recovery task stated in Section~\ref{sec:recovery_task}.
Over the $N_\text{tot}$ episodes run for a given controller, let $n$ denote the number that successfully reach the recovered set $\mathcal{R}$ \eqref{eq:recovery_set} before truncation, and let $T_i$ be the time $T$ at which the $i$-th successful episode reaches $\mathcal{R}$. The overall \emph{recovery rate} ($\mathrm{RR}$) and overall mean \emph{recovery time} ($\mathrm{RT}$) are then
\begin{equation}
    \label{eq:recovery_metrics}
    \mathrm{RR} = \frac{n}{N_\text{tot}}, \qquad
    \mathrm{RT} = \frac{1}{n} \sum_{i=1}^{n} T_i.
\end{equation}

Because the bounds of the initial-state domain are chosen empirically, these metrics are not intrinsically meaningful; rather, they are useful for comparing controllers. We therefore measure efficacy relative to the Stanley baseline controller, partitioning the initial states into those from which the baseline recovers, which we term the Stanley-recoverable set $\mathcal{SR}$, and the remaining states from which it cannot, the Stanley-unrecoverable set $\mathcal{SU}$, with $|\mathcal{SR}| = 38$ and $|\mathcal{SU}| = 256$. Our primary efficacy metric, the \emph{recovery improvement} ($\mathrm{RI}$), is the percentage improvement in total recoveries over that baseline,
\begin{equation}
    \label{eq:stanley_improvement}
    \mathrm{RI} = \frac{n - |\mathcal{SR}|}{|\mathcal{SR}|} \times 100\%.
\end{equation}
On the $\mathcal{SR}$ and $\mathcal{SU}$ subsets we also compare the recovery times $\mathrm{RT}_\text{SR}$ and $\mathrm{RT}_\text{SU}$, and the numbers of recovered states $n_\text{SR}$ and $n_\text{SU}$, out of a possible $38$ and $256$. This enables a comparison on the ``easier'' Stanley-recoverable states and the ``harder'' Stanley-unrecoverable states.

All these quantitative performance metrics are compiled in Table~\ref{tab:recovery-comparison}. We also compare controllers graphically in relation to the $\beta$--$\dot\Psi$ phase plane of Section~\ref{sec:recovery_task}, which captures the difficulty of a state, as shown in Fig.~\ref{fig:recovery_sim_comparison}. This is done by binning the $294$ evaluation episodes by $(\beta_0, \dot\Psi_0)$. This view displays the same overall $\mathrm{RR}$, resolved over the phase plane to emphasize behaviour in relation to drifting dynamics.

\begin{table}[!tb]
    \centering
    \footnotesize
    \renewcommand{\arraystretch}{1.15}
    \setlength{\tabcolsep}{4pt}
    \caption{Evaluation Grid Over the Four Initial-State Variables}
    \label{tab:recovery_eval_grid}
    \begin{tabular}{|c|c|c|c|}
        \hline
        \textbf{Variable} & \textbf{Symbol} & \textbf{Range} & \textbf{Points} \\
        \hline
        Sideslip angle (rad)   & $\beta_0$    & $\pm 1.39$  & 7 \\
        Yaw rate (rad/s)       & $\dot\Psi_0$ & $\pm 13$    & 7 \\
        Velocity (m/s)         & $v_0$        & $[4,\ 5]$   & 2 \\
        Heading error (rad)    & $e_{\Psi,0}$ & $\pm 0.17$  & 3 \\
        \hline
    \end{tabular}

    \vspace{2pt}
    \parbox{\columnwidth}{\scriptsize The grid spans $\pm 79.6^\circ$ of
    sideslip and $\pm 744.8^\circ$/s of yaw rate, far exceeding the
    $\pm 20^\circ$ and $\pm 45^\circ$/s of the training ranges in
    Table~\ref{tab:recovery_init_ranges}. Heading error is $\pm 9.7^\circ$.}
\end{table}

\subsection{Simulation Results}
\label{sec:sim_comparison}

Table~\ref{tab:recovery-comparison} quantitatively compares our approach against the baselines. Stanley recovers from $38$ of the $294$ initial states ($\mathrm{RR} = 12.9\%$), the reference against which $\mathrm{RI}$ is measured. The ST-MPC improves on Stanley by $\mathrm{RI} = 13.2\%$, while every learned controller improves on both classical baselines by an order of magnitude more. The transfer policy $\pi_{\text{race}\to\text{rec}}$ reaches $\mathrm{RI} = 186.8\%$, the best of all methods, as well as the highest recovery count on each subset: $37$ of the $38$ states Stanley recovers, and $72$ of the $256$ it cannot. Recovery times can only be fairly compared on states that both controllers recover, because $\mathrm{RT}$ averages over successful episodes only: a controller that recovers from more of the hard states carries their longer recovery times in its mean. On the common set $\mathcal{SR}$, the transfer policy is more than twice as fast, at $\mathrm{RT}_\text{SR} = 0.347$~s against Stanley's $0.789$~s. The difference in efficacy is also visible graphically in Fig.~\ref{fig:recovery_sim_comparison}(a) versus Fig.~\ref{fig:recovery_sim_comparison}(d). Stanley succeeds almost only in the bottom-left and top-right quadrants, where the yaw rate is already rotating the vehicle out of its sideslip and the state is progressing towards recovery unaided.

\subsection{Ablation}
\label{sec:ablation}

Our ablation addresses two questions: is racing pretraining the key component behind the superior performance of $\pi_{\text{race}\to\text{rec}}$, and, conversely, can the pure-recovery policy $\pi_{\text{rec}}$ close the gap by other means? Both are answered by the ablations block of Table~\ref{tab:recovery-comparison}.

\begin{figure*}[!tb]
  \centering
  \includegraphics{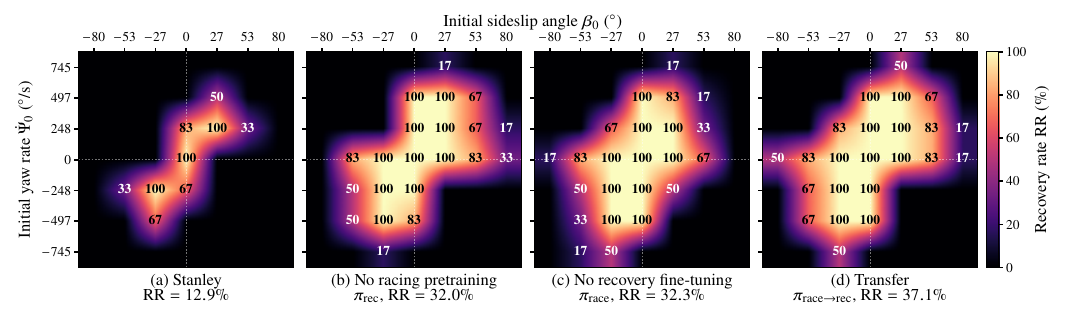}
  \caption[Recovery rate heatmaps across initial states]{Recovery rate
  $\mathrm{RR}$ across the initial states of the nominal evaluation grid of
  Table~\ref{tab:recovery_eval_grid}. Each cell is the percentage of its six
  $(v_0, e_{\Psi,0})$ episodes that recover, so it takes one of seven values;
  cells with no recoveries are left unlabelled, and the dashed lines mark
  $\beta_0 = 0$ and $\dot\Psi_0 = 0$. Each panel is one row of
  Table~\ref{tab:recovery-comparison}, named and marked (a)--(d) as there, and
  its $\mathrm{RR}$ is that row's. The Stanley baseline performs
  poorest, the policies without racing pretraining and without recovery
  fine-tuning similarly well, and the transfer policy best.}
  \label{fig:recovery_sim_comparison}
\end{figure*}

The first two rows address the first question by removing each of the two ingredients in turn: the racing policy $\pi_{\text{race}}$ is evaluated on the recovery task without recovery fine-tuning, and $\pi_{\text{rec}}$ is trained on the recovery task without racing pretraining. Without ever training on recovery, $\pi_{\text{race}}$ already outperforms every classical baseline, at $\mathrm{RI} = 150.0\%$. Fig.~\ref{fig:recovery_sim_comparison}(c) shows where that comes from: it recovers in the challenging top-left and bottom-right quadrants, where the yaw rate is driving the vehicle towards ever larger sideslip. That is exactly the state the racing policy encounters when handling the vehicle at the limit, and on the straight path it applies the drift-exit maneuver it learned while racing. Conversely, $\pi_{\text{rec}}$ struggles in exactly those quadrants (Fig.~\ref{fig:recovery_sim_comparison}(b)) and reaches $\mathrm{RI} = 147.4\%$, level with $\pi_{\text{race}}$ ($94$ recoveries against $95$) despite being trained on the recovery task itself. Racing pretraining alone is thus worth as much as recovery training alone, and the two combined, in the transfer policy, recover from $109$ states. We conclude that racing pretraining is the essential element behind the superior recovery rate, our primary objective.

The last two rows address the second question: whether breadth of experience alone, rather than racing specifically, is what improves performance. Each trains $\pi_{\text{rec}}$ from scratch, still without racing pretraining, on a wider distribution of initial states: the first samples larger $(\beta, \dot\Psi)$ initial ranges directly, and the second reaches the same ranges by curriculum learning~\cite{bengioCurriculumLearning2009}, widening them one stage at a time as the success rate over a rolling window passes a threshold. Neither improves even on $\pi_{\text{rec}}$, at $\mathrm{RI} = 128.9\%$ and $142.1\%$, let alone on the transfer policy, so wider exposure to initial states is no substitute for racing pretraining.

On the secondary objective of recovery time, however, $\pi_{\text{rec}}$ recovers faster than both policies trained on racing, at $\mathrm{RT} = 0.450$~s against $0.740$~s for the transfer policy and $1.227$~s for $\pi_{\text{race}}$. Part of this gap is the bias of the overall $\mathrm{RT}$ noted in Section~\ref{sec:sim_comparison}. On $\mathcal{SR}$, where the transfer policy and $\pi_{\text{rec}}$ recover almost the same states ($37$ and $36$), their times are matched, at $0.347$~s and $0.352$~s; the gap opens on $\mathcal{SU}$, $0.942$~s against $0.511$~s, where the transfer policy recovers $72$ states to $\pi_{\text{rec}}$'s $58$. The racing policy, never trained for minimum-time recovery, is slower even on $\mathcal{SR}$, at $0.969$~s, so it is the fine-tuning that makes the transfer policy fast. The wider-range variant of $\pi_{\text{rec}}$ has the lowest recovery times in the table, overall and on both subsets, though on $\mathcal{SR}$ its lead over the transfer policy is only $0.012$~s. The curriculum variant is the one exception to the speed of recovery-only training: its $\mathrm{RT}$ of $1.375$~s is inflated by a long tail of slow recoveries, its median being $0.455$~s. A trade-off therefore remains, though a smaller one than the overall $\mathrm{RT}$ suggests: learning recovery directly, rather than transferring from racing, achieves a poorer recovery rate but a modestly improved recovery time. This is likely because recovery-only training specializes for the straight-line minimum-time goal, whereas the racing and transfer policies maintain racing-style drifting behaviour that costs time but improves efficacy in more challenging scenarios. In practice, the choice would depend on the application and which objective has higher priority.

Overall, Fig.~\ref{fig:recovery_sim_comparison} and Table~\ref{tab:recovery-comparison} show that learning to race first and then fine-tuning on the recovery task outperforms every alternative that tackles the recovery problem directly.

\begin{table*}[!tb]
    \centering
    \footnotesize
    \renewcommand{\arraystretch}{1.15}
    \caption{Recovery Performance and Ablation on the Nominal Evaluation Grid}
    \label{tab:recovery-comparison}
    \begin{tabular}{|l|c|c|c|c|c|c|c|c|c|}
        \hline
        & & \multicolumn{4}{c|}{\textbf{Overall}}
        & \multicolumn{2}{c|}{\textbf{Stanley-rec.\ ($\mathcal{SR}$)}}
        & \multicolumn{2}{c|}{\textbf{Stanley-unrec.\ ($\mathcal{SU}$)}} \\
        \hline
        \textbf{Controller} & \textbf{Fig.~\ref{fig:recovery_sim_comparison}}
            & $\mathrm{RI}\,\uparrow$ & $\mathrm{RR}\,\uparrow$ & $n\,\uparrow$ & $\mathrm{RT}\,\downarrow$
            & $n_\text{SR}\,\uparrow$ & $\mathrm{RT}_\text{SR}\,\downarrow$
            & $n_\text{SU}\,\uparrow$ & $\mathrm{RT}_\text{SU}\,\downarrow$ \\
            & & (\%) & (\%) & (of 294) & (s)
            & (of 38) & (s) & (of 256) & (s) \\
        \hline
        \multicolumn{10}{|l|}{\textit{Baseline controllers}} \\
        \hline
        Stanley & (a) & --- & 12.9 & 38 & 0.789 & --- & 0.789 & --- & --- \\
        ST-MPC  & & $+13.2$ & 14.6 & 43 & 0.813 & 34 & 0.774 & 9 & 0.961 \\
        \hline
        \multicolumn{10}{|l|}{\textit{Proposed controller}} \\
        \hline
        Transfer, $\pi_{\text{race}\to\text{rec}}$ & (d)
            & $\mathbf{+186.8}$ & $\mathbf{37.1}$ & $\mathbf{109}$ & 0.740
            & $\mathbf{37}$ & 0.347 & $\mathbf{72}$ & 0.942 \\
        \hline
        \multicolumn{10}{|l|}{\textit{Ablations}} \\
        \hline
        No recovery fine-tuning, $\pi_{\text{race}}$ & (c)
            & $+150.0$ & 32.3 & 95 & 1.227 & 36 & 0.969 & 59 & 1.384 \\
        No racing pretraining, $\pi_{\text{rec}}$ & (b)
            & $+147.4$ & 32.0 & 94 & 0.450
            & 36 & 0.352 & 58 & 0.511 \\
        $\pi_{\text{rec}}$, wider initial ranges &
            & $+128.9$ & 29.6 & 87 & $\mathbf{0.391}$
            & 36 & $\mathbf{0.335}$ & 51 & $\mathbf{0.431}$ \\
        $\pi_{\text{rec}}$, curriculum &
            & $+142.1$ & 31.3 & 92 & 1.375 & 36 & 0.873 & 56 & 1.698 \\
        \hline
    \end{tabular}

\end{table*}

\section{Hardware Validation}
\label{sec:hardware_validation}

Deploying the recovery controller of Section~\ref{sec:transfer} on a physical vehicle is the most demanding validation of the approach, as physical testing exposes limitations that simulation does not: vehicle parameter mismatch, actuator delays, and state observation noise. The results of Section~\ref{sec:sim_results} use the nominal vehicle parameters of Section~\ref{sec:racing_policy}; here they are replaced by parameters identified on the physical vehicle as in Section~\ref{sec:sim2real-transfer}, and the controller is retrained on them. We describe the testbed and its deployment in Section~\ref{sec:deployment}, then the real-vehicle experiments in Section~\ref{sec:hardware_results}.

\subsection{Testbed and Deployment}
\label{sec:deployment}

\begin{figure}[t]
  \centering
  \includegraphics[width=\columnwidth]{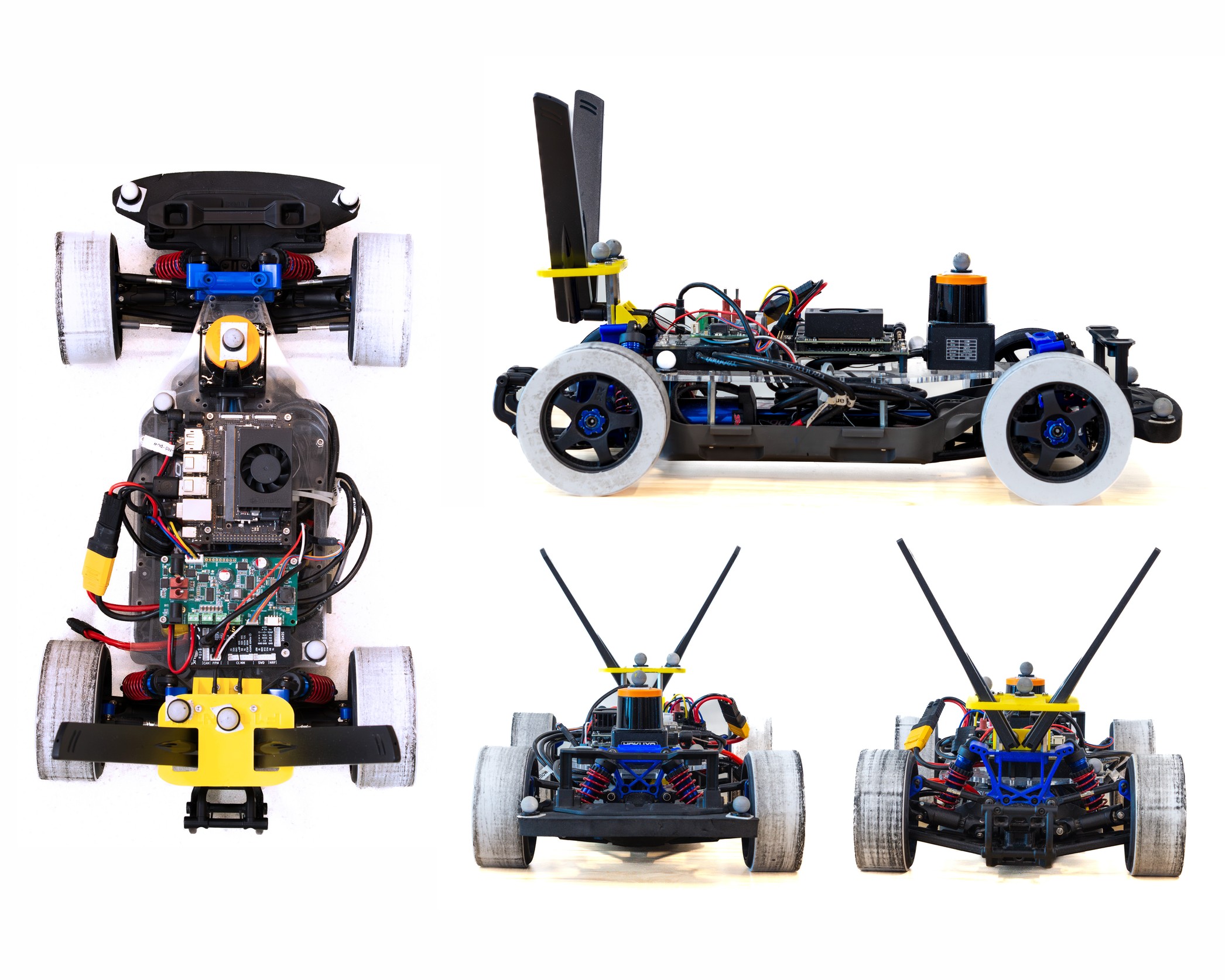}
  \caption[Modified F1TENTH testbed]{The modified F1TENTH hardware platform from varying angles, showing custom tire covers, lower ride height, harder shock absorbers, and stock F1TENTH hardware components.}
  \label{fig:f1tenth}
\end{figure}

We choose the F1TENTH 1:10-scale autonomous racecar platform as our testbed~\cite{okellyF110OpenSource2019},\footnote{\raggedright F1TENTH build docs: \url{https://f1tenth.readthedocs.io/}} because it enables agile vehicle control without the risks associated with full-size testing. Two modifications adapt it to our purposes. First, harder shock absorbers, a lowered ride height and a faster steering servo make the vehicle physically capable of executing the drift maneuvers we require. Second, we find a tire--road combination whose friction coefficient is sufficiently high that the vehicle is still controllable, but sufficiently low that recovering control is a challenge at the relatively low speeds at which we perform experiments: 3D-printed tire covers on a smooth epoxy floor. The resulting testbed is shown in Fig.~\ref{fig:f1tenth}. 

The simulation action vector~\eqref{eq:racing_action} matches the steering-angle and speed commands accepted by the F1TENTH's VESC motor controller, so the policy's commands are passed to the vehicle unchanged, leaving the actuator response as the gap to model, which \eqref{eq:first_order_lag} does.

The observation \eqref{eq:racing_obs} is recoverable from a combination of vehicle sensors and a motion-capture system. The average wheel speed $\omega_{\text{avg}}$ is reported by the motor controller, and the lookahead curvatures $\boldsymbol{\kappa}$ are all zero on the straight reference path. All remaining quantities are recovered from a motion-capture system running at 200~Hz using a virtual reference line, with the same track width of 2.2~m used in simulation. As in simulation, the controller runs at 100~Hz. 
One quantity is not recoverable: the steering servo angle, as the device does not have an encoder to publish position feedback.

The identified parameters, actuator dynamics and randomization of Section~\ref{sec:sim2real-transfer} are implemented in our Gym-Khana simulator. As such, both the racing task and the recovery task are executed using realistic parameters. This way, the dynamics knowledge is that of the real vehicle. The policies are retrained from scratch with the realistic model: the racing policy $\pi_{\text{race}}$ is retrained as in Section~\ref{sec:racing_policy} and transferred to the recovery task as in Section~\ref{sec:transfer},
yielding the $\pi_{\text{race}\to\text{rec}}$ deployed and evaluated in Section~\ref{sec:hardware_results}.

To generate initial states, we teleoperate the car into the test area forcing non-recovered states, $\mathbf{z}_0 \in \mathcal{U}$. We record the control inputs and adjust their magnitudes until they produce feasible and meaningful initial states. During experiments, we replay the pre-recorded commands from the same starting position on every run.
Because the physical state cannot be reset precisely, the initial steering angle may not be exactly $\delta_0=0$ as in simulation, and a nonzero lateral deviation may already be present.
As such, under realistic conditions, even though the same set of pre-recorded control commands is used, the resulting initial state, $\mathbf{z}_0$, may slightly vary between runs. Therefore, the experiment is repeated 30 times to characterize this variability.

On the vehicle, a ROS2 package assembles the observation from the motion-capture and sensor streams, runs the policy at 100~Hz, and tests membership of the recovered set $\mathcal{R}$ of \eqref{eq:recovery_set} online.\footnote{\raggedright Recovery ROS2 package: \url{https://github.com/TeoIlie/recovery_controller}} The rates $\dot\beta$ and $\ddot\Psi$ are obtained by numerical differentiation of the motion-capture signals; the one quantity not measured on our platform is the steering angle, so the $\delta$ threshold is evaluated on the modelled value from the first-order lag \eqref{eq:first_order_lag}. 
Because recovery is defined here as membership of a region,
$\mathbf{z}_T \in \mathcal{R}$, when recovered the membership check must be able to hand back control to the default vehicle controller. In our case, recovery or failure commands a full vehicle stop.

Of the two baselines evaluated in simulation, only Stanley was ported to the physical vehicle, and is therefore used as the baseline for hardware validation. The full 294-state evaluation grid is impractical to reproduce on the physical vehicle, so we select three initial states of increasing difficulty for hardware evaluation. The states were chosen to produce a mixture of recoveries and failures rather than being consistently recovered or failed by both controllers.

\subsection{Real-Vehicle Experiments}

\label{sec:hardware_results}

Each run has three phases in the motion-capture space: a pre-recorded sequence of control inputs induces the initial state $\mathbf{z}_0 \in \mathcal{U}$, the controller then takes full authority and attempts to bring the vehicle into $\mathcal{R}$, and the car brakes on recovery or on reaching the end of the run area.

The three selected initial states are summarized in Table~\ref{tab:real_init_states}, which reports their median values and interquartile ranges (IQR) across the experimental runs.
The easy state sits near the stable equilibrium. The medium state lies in the challenging top-left quadrant of Fig.~\ref{fig:recovery_sim_comparison}, its yaw rate rotating the vehicle towards ever larger sideslip, and the hard state in the corresponding bottom-right quadrant, farther from the equilibrium and with a $-36^\circ$ heading error on top.

\begin{table}[!tb]
    \centering
    \scriptsize
    \renewcommand{\arraystretch}{1.2}
    \setlength{\tabcolsep}{3pt}
    \caption{The Three Real-Vehicle Initial States}
    \label{tab:real_init_states}
    \begin{tabular}{|c|c|c|c|}
        \hline
        \textbf{Symbol} & \textbf{Easy} & \textbf{Medium} & \textbf{Hard} \\
        \hline
        $\beta_0$ ($^\circ$)    & $-6$ ($-9$, $-3$)   & $-19$ ($-24$, $-13$) & $28$ ($25$, $34$) \\
        $\dot\Psi_0$ ($^\circ$/s) & $45$ ($14$, $100$)  & $92$ ($39$, $154$)   & $-120$ ($-176$, $-70$) \\
        $v_0$ (m/s)         & $3.9$ ($3.8$, $4.1$) & $3.5$ ($3.3$, $3.6$) & $3.2$ ($3.1$, $3.4$) \\
        $e_{\Psi,0}$ ($^\circ$) & $-16$ ($-19$, $-14$) & $0$ ($-2$, $2$)      & $-36$ ($-37$, $-35$) \\
        $\delta_0$ ($^\circ$)   & $15$ ($15$, $15$)   & $0$ ($-12$, $8$)     & $0$ ($0$, $0$) \\
        \hline
    \end{tabular}

    \vspace{2pt}
    \parbox{\columnwidth}{\scriptsize Median (IQR) of the initial sideslip,
    yaw rate, velocity, heading error and steering angle, logged over all runs
    of both controllers, excluding four runs with
    spurious yaw rates $|\dot\Psi_0| > 10$~rad/s caused by numerical
    differentiation of the raw motion-capture pose. Those runs are included in
    all recovery metrics.}
\end{table}

\begin{figure}[!tb]
  \centering
  \subfloat[Recovery rate $\mathrm{RR}$ per initial state.]{%
    \includegraphics{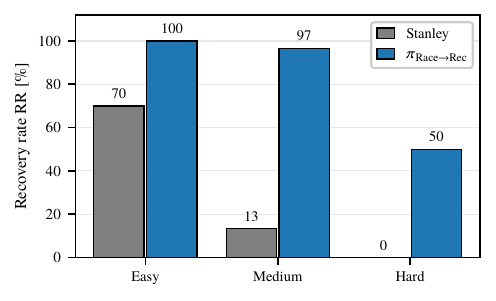}%
    \label{fig:real_rr}}
  \\
  \subfloat[Outcome of every run per initial state.]{%
    \includegraphics{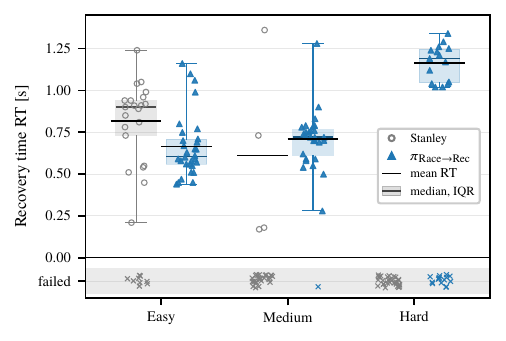}%
    \label{fig:real_rt}}
  \caption[Real-vehicle recovery results]{Real-vehicle recovery results for
  the Stanley baseline and the transfer policy $\pi_{\text{race}\to\text{rec}}$,
  a qualitative view of the 179 runs of Table~\ref{tab:real_results}.
  (a)~Recovery rate. (b)~Recovered runs are plotted at their recovery time,
  with the mean $\mathrm{RT}$ marked by a black bar; failed runs are shown as
  crosses in the band below. Box plots show $\mathrm{RT}$ distributions, and
  are omitted on the Stanley medium and hard states where there are too few
  successful runs for a meaningful measurement.}
  \label{fig:real_results}
\end{figure}

We run both controllers 30 times on each state, for $3 \times 2 \times 30 = 180$ runs; one $\pi_{\text{race}\to\text{rec}}$ run on the medium state was found corrupted during the data-processing stage, leaving 179 valid runs. The results are given in Table~\ref{tab:real_results} and shown qualitatively in Fig.~\ref{fig:real_results}, which compares the recovery rates (Fig.~\ref{fig:real_results}(a)) and plots the outcome of every run (Fig.~\ref{fig:real_results}(b)).

The policy $\pi_{\text{race}\to\text{rec}}$ recovers far more often than Stanley on every state, for example $100\%$ against $70\%$ on the easy state
and $50\%$ against $0\%$ on the hard state, where Stanley fails on every run. On the easy state, $\pi_{\text{race}\to\text{rec}}$ is also faster
and more consistent.
That consistency holds on the harder states, where its IQR is $0.15$~s wide on the medium state and $0.20$~s on the hard one. Stanley's medium-state $\mathrm{RT}$ is lower, but the two are not comparable: the metric is defined over recovered runs only, and Stanley recovers $13.3\%$ of them, necessarily the easiest draws of the state. This is why Section~\ref{sec:sim_comparison} compares recovery times on the common set $\mathcal{SR}$.

The gap widens as the state becomes more challenging. While the baseline still copes in easier states, it never recovers in the hardest state. The advantage therefore survives sim-to-real transfer precisely where the controller is most needed: where agile, drift-style control succeeds but conservative path tracking fails.

\begin{table}[!tb]
    \centering
    \footnotesize
    \renewcommand{\arraystretch}{1.15}
    \setlength{\tabcolsep}{4pt}
    \caption{Real-Vehicle Recovery Performance per Initial State}
    \label{tab:real_results}
    \begin{tabular}{|l|c|c|c|c|c|}
        \hline
        \textbf{Controller}
            & $N$ & $n\,\uparrow$ & $\mathrm{RR}\,\uparrow$ & $\mathrm{RT}\,\downarrow$
            & $\mathrm{RT}$ median (IQR)$\,\downarrow$ \\
            & & & (\%) & (s) & (s) \\
        \hline
        \multicolumn{6}{|l|}{\textit{Easy state}} \\
        \hline
        Stanley & 30 & 21 & 70.0 & 0.815 & 0.900 (0.730, 0.939) \\
        $\pi_{\text{race}\to\text{rec}}$ & 30 & $\mathbf{30}$ & $\mathbf{100.0}$ & $\mathbf{0.665}$
            & $\mathbf{0.605}$ (0.560, 0.707) \\
        \hline
        \multicolumn{6}{|l|}{\textit{Medium state}} \\
        \hline
        Stanley & 30 & 4 & 13.3 & $\mathbf{0.610}$ & $\mathbf{0.455}$ (0.178, 0.888) \\
        $\pi_{\text{race}\to\text{rec}}$ & 29 & $\mathbf{28}$ & $\mathbf{96.6}$ & 0.709
            & 0.720 (0.612, 0.765) \\
        \hline
        \multicolumn{6}{|l|}{\textit{Hard state}} \\
        \hline
        Stanley & 30 & 0 & 0.0 & --- & --- \\
        $\pi_{\text{race}\to\text{rec}}$ & 30 & $\mathbf{15}$ & $\mathbf{50.0}$ & $\mathbf{1.164}$
            & $\mathbf{1.190}$ (1.045, 1.245) \\
        \hline
    \end{tabular}

    \vspace{2pt}
    \parbox{\columnwidth}{\scriptsize Over $N$ runs of which $n$ recovered.
    $\mathrm{RT}$, and the median and IQR of the recovery time, are computed
    over the $n$ recovered runs only, so Stanley's medium-state times are not
    comparable with the policy's.}
\end{table}

\section{Conclusion}
\label{sec:conclusion}
In this paper, we explored the use of reinforcement learning (RL) to design a controller that recovers a vehicle from loss of control at the limit of handling. We proposed a racing-to-recovery transfer method, in which a policy is first trained to race on a track, where drifting emerges on its own as the time-optimal way to take a corner, and is then adapted to the recovery task by continuing training under a minimum-time recovery reward. The two tasks were formulated to share a single vehicle model, observation space, and action space, so that only the reward changes between them. We grounded the notion of a ``recovered'' state in the $\beta$--$\dot\Psi$ phase plane rather than in an empirical, scenario-specific criterion, and defined a set of metrics that make recovery controllers comparable on a common domain of loss-of-control states. Both the racing and recovery policies were trained in Gym-Khana, a simulator we built around a drift-capable single-track model, and the resulting controller was deployed on a physical F1TENTH vehicle via a sim-to-real pipeline of system identification, actuator-lag modelling, and domain randomization.

We showed that this transfer approach outperforms classical control baselines and a recovery-only policy trained from scratch. In simulation, over a grid of 294 initial states more extreme than any seen in training, the transferred policy improved recoveries by $186.8\%$ over a Stanley path tracker, exceeding both a nonlinear MPC and the recovery-only policy. Our ablation showed that this advantage comes specifically from racing pretraining rather than from broader exposure to initial states alone. Note that our claim is not that racing pretraining is the only route to a competent recovery controller: a recovery-only policy trained directly on a wider initial-state distribution recovered slightly faster on the states it could already handle, even though it recovered from fewer of them overall. Instead, our results support the claim that transferring from racing is what lets a policy generalize to the broadest range of loss-of-control states, and that this advantage survives sim-to-real transfer. When deployed on hardware, the transferred policy outperformed the Stanley baseline on every one of three real initial states over around 180 runs, including the hardest state, from which the baseline never recovered.

Although the proposed framework demonstrates promising results, several directions remain for future research. First, future work will extend validation beyond a straight reference path to curved paths and general driving lanes, where the recovery task must additionally account for path curvature. Second, while recovery was evaluated using external motion-capture state estimation, future studies may explore recovery under onboard, noisier state estimation, closer to what a deployed ADAS would have available. Another important avenue is a formal statistical characterization of the drifting maneuvers the transferred policy exploits during recovery, to make precise what racing pretraining contributes beyond the empirical gains shown here. Finally, extending real-vehicle validation to a broader set of initial states, and studying the efficacy-versus-speed trade-off we observed between transfer and recovery-only training under different mission priorities, may further improve the framework's applicability to real-world autonomous driving systems.

\bibliographystyle{IEEEtran}
\bibliography{refs}

\begin{IEEEbiography}[{\includegraphics[width=1in,height=1in,clip,keepaspectratio]{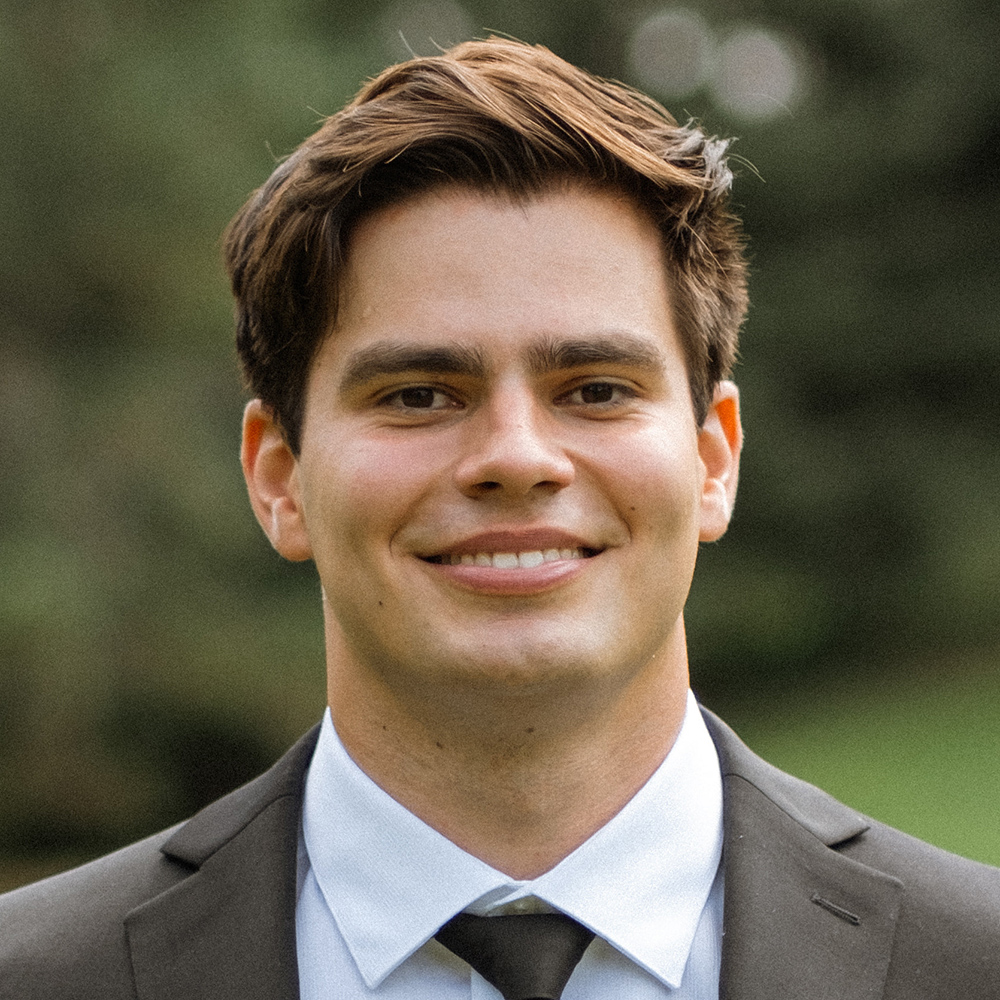}}]{Teodor Ilie}
(Member, IEEE) received the B.Sc. (Hons) and M.Sc. degrees in Computer Science from Queen's University, Kingston, ON, Canada, in 2022 and 2026, respectively. He is currently with MacLean Engineering, Barrie, ON, Canada, working on vehicle autonomy. His research interests include learning-based control, vehicle dynamics, autonomous racing, and safety.
\end{IEEEbiography}

\begin{IEEEbiography}[{\includegraphics[width=1in,height=1in,clip,keepaspectratio]{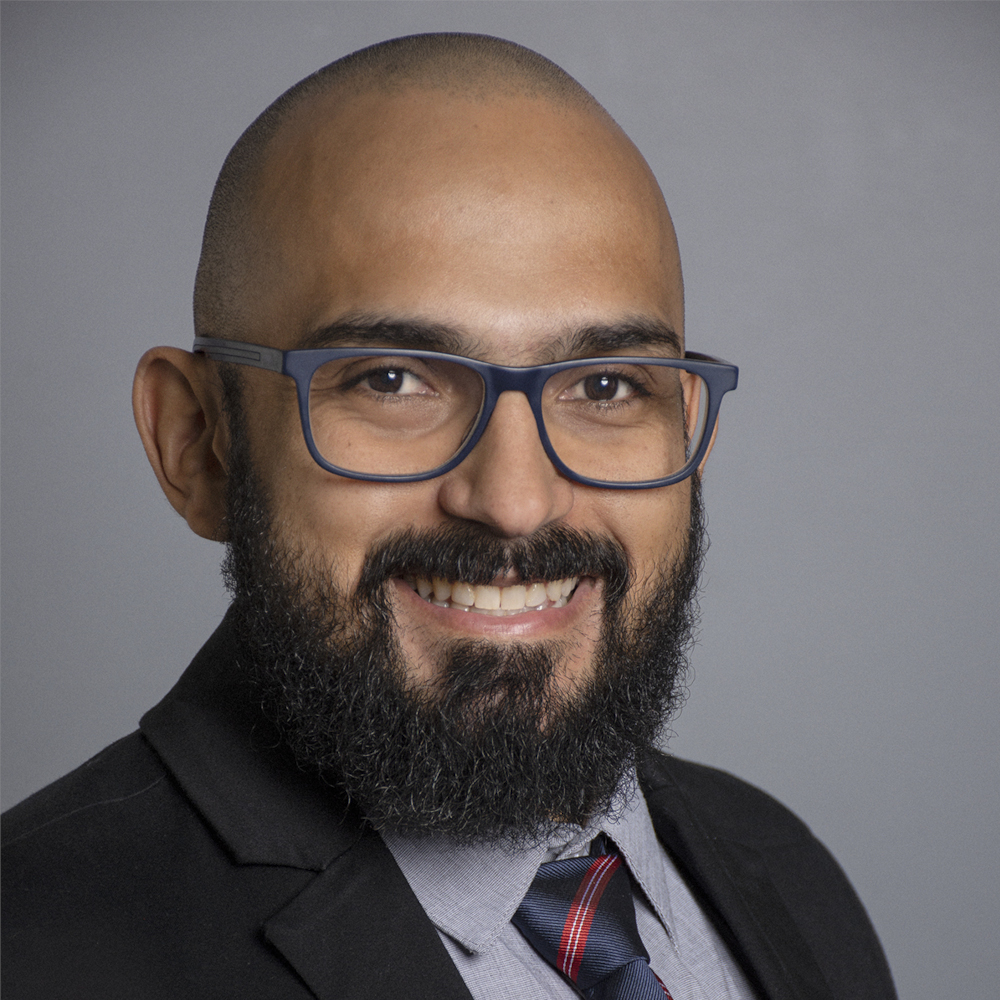}}]{Kleber Cabral}
(Member, IEEE) is a Postdoctoral Fellow and Adjunct Professor at Queen’s University. He holds a Ph.D. in Electrical and Computer Engineering from RMC, an M.Sc. from the Aeronautics Institute of Technology (ITA, Brazil), and a B.Sc. in Computer Engineering from the Federal University of Goias (UFG, Brazil). His research interests include multi-robot systems, UAV/UGV autonomy, task and mission assignment, geometric and decentralized control, and deep reinforcement learning for multi-agent coordination. His recent work spans swarm control, mechanism design for robot cooperation, and learning-based navigation.
\end{IEEEbiography}

\begin{IEEEbiography}[{\includegraphics[width=1in,height=1in,clip,keepaspectratio]{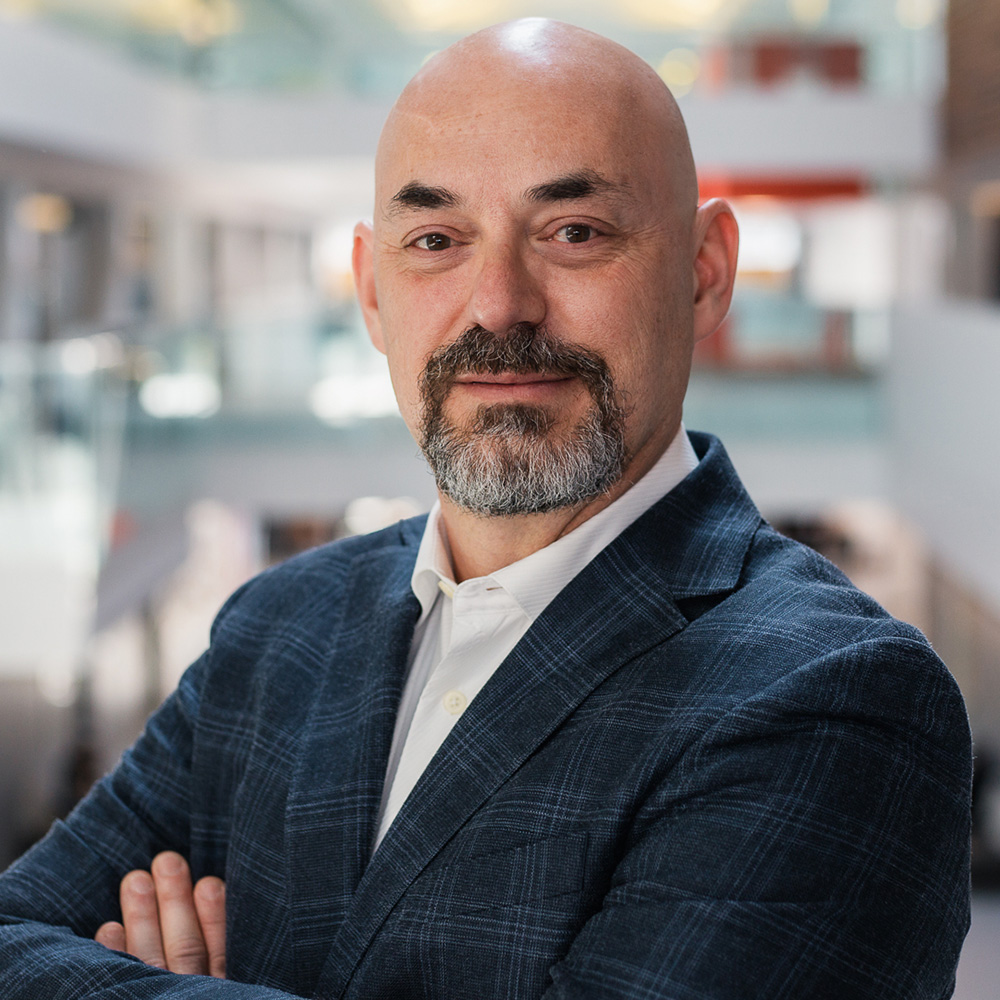}}]{Sidney Givigi}
(Senior Member, IEEE) received the Ph.D. degree in electrical and computer engineering from Carleton University, Ottawa, ON, Canada, in 2009. He is currently a Professor with the School of Computing, Queen’s University, Canada. He was with Carleton University and the Royal Military College of Canada. His research interests are mainly focused on autonomous systems and robotics, especially the decentralized control of multiple vehicles, learning and adaptation of robots, and modeling of complex systems with game theory.
\end{IEEEbiography}

\end{document}